\documentclass[11pt]{article}

\usepackage[preprint]{acl}

\usepackage{times}
\usepackage{latexsym}

\usepackage[T1]{fontenc}

\usepackage[utf8]{inputenc}

\usepackage{microtype}
\usepackage{amsmath}
\usepackage{amssymb}
\usepackage{algorithm}
\usepackage{algpseudocode}
\usepackage{tcolorbox}
\usepackage{multirow}
\usepackage{booktabs}
\usepackage{array}
\usepackage{subcaption}
\usepackage{pifont}
\usepackage[dvipsnames]{xcolor}
\usepackage{colortbl}   
\usepackage{longtable}
\usepackage{soul}
\usepackage{inconsolata}

\usepackage{graphicx}

\definecolor{att0}{RGB}{255,255,255} 
\definecolor{att1}{RGB}{245,248,252} 
\definecolor{att2}{RGB}{232,240,250} 
\definecolor{att3}{RGB}{210,225,242} 
\definecolor{att4}{RGB}{185,205,230}
\definecolor{att5}{RGB}{160,185,215}
\definecolor{att6}{RGB}{130,160,195}

\newcommand{\att}[2]{\sethlcolor{#1}\hl{#2}}

\title{KV-Embedding: Training-free Text Embedding via Internal KV Re-routing in Decoder-only LLMs}

\author{Yixuan Tang \and Yi Yang \\
The Hong Kong University of Science and Technology\\
\texttt{ytangch@connect.ust.hk, imyiyang@ust.hk}
}

\begin{document}
\maketitle
\begin{abstract}
  While LLMs are powerful embedding backbones, their application in training-free settings faces two structural challenges: causal attention restricts early tokens from accessing subsequent context, and the next-token prediction objective biases representations toward generation rather than semantic compression. To address these limitations, we propose KV-Embedding, a framework that activates the latent representation power of frozen LLMs. Our method leverages the observation that the key-value (KV) states of the final token at each layer encode a compressed view of the sequence. By re-routing these states as a prepended prefix, we enable all tokens to access sequence-level context within a single forward pass. To ensure model-agnostic applicability, we introduce an automated layer selection strategy based on intrinsic dimensionality. Evaluations on MTEB across Qwen, Mistral, and Llama backbones show that KV-Embedding outperforms existing training-free baselines by up to 10\%, while maintaining robust performance on sequences up to 4,096 tokens. These results demonstrate that internal state manipulation offers an efficient alternative to input modification, and we hope this work encourages further exploration of LLM internals for representation learning.
\end{abstract}

\section{Introduction}

\begin{figure*}
  \centering
  \includegraphics[width=\textwidth]{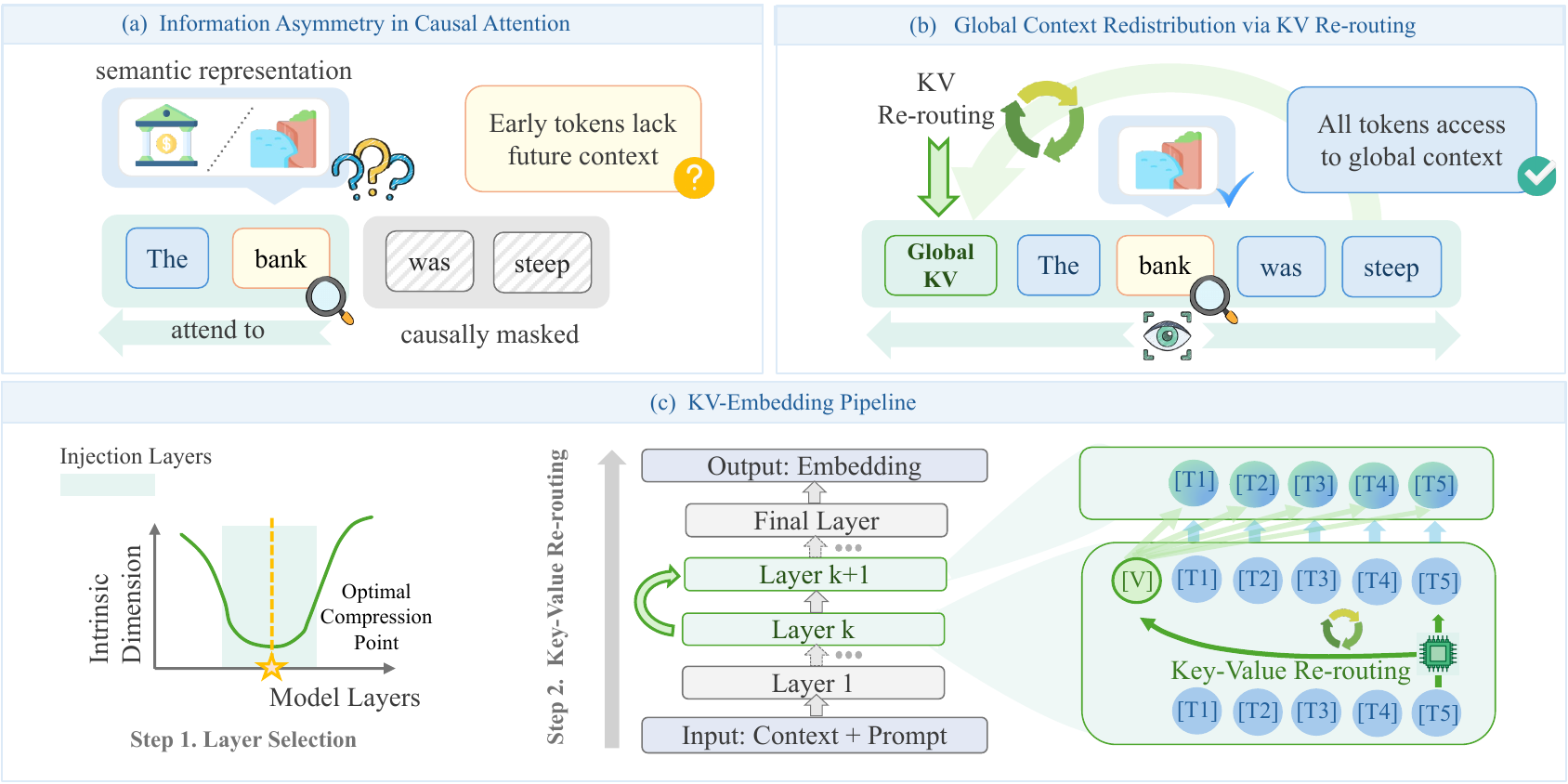}
  \caption{Overview of KV-Embedding. (a) Standard causal attention creates information asymmetry, as tokens only attend to preceding context. (b) KV-Embedding re-routes the last KV pair as a global prefix, enabling sequence-wide context access within one forward pass. (c) The pipeline identifies optimal re-routing anchors by locating layers with minimum intrinsic dimensionality, ensuring model-agnostic stability without manual tuning.}
  \label{fig:method}
\end{figure*}

Text embeddings power a wide range of NLP applications, from semantic search to retrieval-augmented generation~\citep{partcirag,songrag}. Recent work has shown that decoder-only LLMs can serve as strong embedding backbones when fine-tuned with contrastive objectives~\citep{wang2024improving, behnamghader2024llm2vec}. However, these methods require substantial computational resources and large-scale datasets for each new backbone, limiting their scalability to rapidly evolving model architectures.

An appealing alternative is training-free embedding extraction: obtaining representations directly from frozen LLMs. This setting requires no optimization and applies immediately to any new model. However, two structural properties of decoder-only LLMs pose significant challenges. First, causal attention restricts each token to attend only to preceding positions, creating an information asymmetry where early tokens remain unaware of subsequent context~\citep{springer2025echo}. Second, the next-token prediction objective biases the final token toward predicting future content rather than distilling the semantic essence of the input~\citep{jiang2024prompteol}. Consequently, standard pooling strategies either average over incomplete representations (mean pooling) or rely on a prediction-biased vector (last-token pooling). For example, consider the phrase ``the bank of the river'': due to the causal mask, the representation of ``bank'' is computed without knowledge of ``river'' and is thus inherently ambiguous. Since mean pooling aggregates all token states equally, this ambiguous representation degrades the quality of the final embedding.

Existing training-free methods address these limitations through various strategies. To mitigate the prediction bias, PromptEOL~\citep{jiang2024prompteol} uses specially designed prompts to guide the final token toward semantic representation. To address information asymmetry, Echo~\citep{springer2025echo} uses input repetition, while token prepending~\citep{fu2025tokenprepending} inserts special tokens, both aiming to provide each token with global context. However, both Echo and token prepending rely on input-level modifications, which introduce their own limitations: input repetition doubles the sequence length, leading to a quadratic increase in attention complexity and potential ``lost-in-the-middle'' effects~\citep{liu-etal-2024-lost}; token prepending relies on special tokens outside the model's vocabulary, leading to unpredictable representations.

We take a different approach. Rather than reconstructing global context through input-level modifications, we observe that the model already computes such global information internally. Specifically, the final token's hidden state at each layer aggregates information from all preceding positions due to the causal attention structure~\citep{radford2019language}. This raises a natural question: \textit{can we directly leverage these internally computed global representations to produce higher-quality text embeddings?}

To this end, we propose \textbf{KV-Embedding}, a training-free framework that re-routes the Key and Value (KV) states of the final token as an internal prefix. As illustrated in Figure~\ref{fig:method}, the final token's KV pair at each layer encodes that layer's compressed view of the sequence~\citep{katharopoulos2020transformers, mor2021transformer}. By prepending this pair to the attention mechanism, we enable all tokens to access a global summary within a single forward pass. We also adopt a prompt-based strategy to mitigate prediction bias in the final token. This design achieves sequence-wide context access without doubling the sequence length or introducing out-of-vocabulary tokens.

We further introduce an intrinsic dimensionality-based layer selection strategy to ensure model-agnostic applicability. Rather than performing re-routing at every layer, we identify layers where representations exhibit maximal compression, corresponding to points where the latent manifold dimensionality is minimized. By targeting these layers as re-routing anchors, we avoid interfering with early-layer feature extraction or late-layer prediction heads, ensuring that the redistributed context aligns with the model's internal geometry. 

Evaluations on MTEB~\citep{muennighoff2023mteb} and the long-context retrieval benchmark LoCoV1~\citep{saad2024locov1} demonstrate the efficacy of KV-Embedding. On MTEB, the proposed framework outperforms existing training-free methods by up to 10\% across three backbones: Qwen3-4B~\citep{yang2025qwen3}, Mistral-7B~\citep{jiang2023mistral}, and Llama-3.1-8B~\citep{dubey2024llama}. On LoCoV1, KV-Embedding maintains robust performance for sequences up to 4,096 tokens, where baseline methods degrade due to context dilution. Further analysis confirms that our method yields a more isotropic embedding space with improved alignment and uniformity.

In summary, KV-Embedding establishes a new state-of-the-art for training-free text embeddings through internal KV re-routing. The framework maintains model-agnostic applicability via automated layer selection based on intrinsic dimensionality, eliminating manual cross-architecture tuning. Extensive evaluations demonstrate that KV-Embedding provides an efficient and scalable solution for leveraging LLMs as text embedding models.

\section{Related Work}
The proposed framework builds upon three research directions: LLM-based text embeddings, training-free representation extraction, and internal state manipulation within the attention mechanism.

\paragraph{LLM-based Text Embeddings}
Text embedding has evolved from encoder-based models to decoder-only LLMs. Early explorations such as SGPT~\citep{muennighoff2022sgpt} utilized position-weighted pooling, while subsequent models like E5-Mistral~\citep{wang2024improving} and GTE~\citep{li2023gte} achieved state-of-the-art results through contrastive fine-tuning on large-scale synthetic datasets. To overcome the inherent limitations of causal masking, LLM2Vec~\citep{behnamghader2024llm2vec} and NV-Embed~\citep{lee2024nvembed} modify the attention mask during training to enable bidirectional context. GritLM~\citep{muennighoff2024gritlm} further unifies generative and embedding capabilities within a single architecture. While these training-based methods yield strong representations, the requirement for extensive computational resources and specialized datasets limits their scalability to the rapidly expanding landscape of LLMs. This motivates the development of training-free approaches that can leverage new models without additional optimization. Our work falls into this category, focusing on unlocking the representational capacity of frozen LLMs through internal state manipulation.

\paragraph{Training-Free LLM Embeddings}
Training-free methods aim to extract high-quality embeddings from frozen LLMs without parameter updates. However, this setting is constrained by the structural properties of decoder-only architectures: the next-token prediction objective biases final states toward future predictions rather than semantic summarization~\citep{jiang2024prompteol}, and causal attention creates an information asymmetry that leaves early tokens unaware of subsequent context~\citep{springer2025echo}. To address the former, PromptEOL~\citep{jiang2024prompteol} and MetaEOL~\citep{lei2024metataskprompting} design specific prompts to guide the model toward semantic compression. To address the latter, Echo~\citep{springer2025echo} repeats the input sequence to simulate bidirectionality, though at the cost of doubling the sequence length. Token Prepending~\citep{fu2025tokenprepending} mitigates this overhead by propagating auxiliary tokens, but relies on out-of-vocabulary tokens that may lead to unpredictable representations. In contrast, our method redistributes internal states within the model's native attention mechanism, enabling sequence-wide context access without modifying the input or introducing external tokens.

\paragraph{KV Manipulation in Attention}
In the Transformer architecture, the interaction between Query, Key, and Value determines the information flow across the sequence~\citep{vaswani2017attention}. Recent interpretability studies suggest that KV pairs function as a form of associative memory, where Keys encode addressing patterns and Values store semantic content~\citep{mor2021transformer}. Research on information flow further indicates that global context accumulates within the KV states of specific anchor tokens during forward propagation~\citep{wang2023labelwords}. This provides a natural entry point for internal state manipulation. Prefix tuning~\citep{li2021prefix} prepends learnable KV pairs as a parameter-efficient fine-tuning strategy, while KV cache compression methods like StreamingLLM~\citep{xiao2024streamingllm} and H2O~\citep{zhang2024h2o} selectively retain entries to optimize inference efficiency. These approaches either require training or target generative throughput. In contrast, KV-Embedding leverages KV manipulation for representation extraction. By re-routing the final token's KV states as a prefix within selected layers, our framework enables early tokens to access a compressed sequence summary, resolving the causal attention limitation within a single forward pass.

\section{Method}

We present KV-Embedding, a training-free framework for extracting high-quality text embeddings from decoder-only LLMs. The pipeline is illustrated in Figure~\ref{fig:method}.

\subsection{Preliminaries}

Given an input sequence $X = (x_1, \dots, x_n)$, a decoder-only LLM produces hidden states $\mathbf{h}_i^{(l)} \in \mathbb{R}^d$ through $L$ sequential layers. For each layer $l$, the hidden state $\mathbf{h}_i^{(l)}$ is computed via multi-head causal self-attention. For simplicity, we describe the single-head case:
\begin{equation}
    \mathbf{h}_i^{(l)} = \text{Attention}\left(\mathbf{q}_i^{(l)}, \mathbf{K}_{\le i}^{(l)}, \mathbf{V}_{\le i}^{(l)}\right)
\end{equation}
where $\mathbf{K}_{\le i}^{(l)} = [\mathbf{k}_1^{(l)}, \dots, \mathbf{k}_i^{(l)}]$ and $\mathbf{V}_{\le i}^{(l)} = [\mathbf{v}_1^{(l)}, \dots, \mathbf{v}_i^{(l)}]$ denote the Key and Value matrices from preceding positions. This causal constraint ensures that each $\mathbf{h}_i$ is computed without visibility into subsequent tokens $x_{j>i}$. To extract a fixed-size embedding $\mathbf{e}$ from a frozen LLM, a pooling function is applied to the final-layer hidden states:
\begin{equation}
    \mathbf{e} = \text{Norm}\left(P\left(\mathbf{h}_1^{(L)}, \dots, \mathbf{h}_n^{(L)}\right)\right)
\end{equation}
where $P(\cdot)$ denotes the pooling strategy and $\text{Norm}(\cdot)$ represents $\ell_2$ normalization.

\subsection{KV-Embedding}
KV-Embedding addresses the structural limitations of causal attention through three components: a compression-oriented prompt, an internal KV re-routing mechanism, and an automated layer selection strategy.

\subsubsection{Compression-Oriented Prompting}

To mitigate the inherent next-token prediction bias, we wrap the input text in a template designed to trigger semantic compression:
\begin{quote}
\textit{``\{Context/Query\}: \{text\}'' Compress the \{Context/Query\} in one word:}
\end{quote}
The prefix is set as ``Context'' for documents and ``Query'' for queries. This template leverages the model's generation capability to guide the final token representation $\mathbf{h}_n$ toward distilling the semantic essence of the input.

\subsubsection{Key-Value Re-Routing}

While prompting shapes the optimization target, it does not resolve the unidirectional information flow: early tokens still cannot access subsequent context. We address this by re-routing the KV states of the final token as a global prefix within each selected layer.

At each layer $l \in \mathcal{L}$, we first compute the Key and Value projections for all positions, yielding the full matrices $\mathbf{K}^{(l)}$ and $\mathbf{V}^{(l)}$. We then extract the KV pair at the final position $(\mathbf{k}_n^{(l)}, \mathbf{v}_n^{(l)})$, which encodes that layer's aggregated view of the sequence due to the causal structure. This pair is prepended to the KV matrices before computing attention:
\begin{equation}
\begin{split}
    \tilde{\mathbf{K}}^{(l)}_{\le i} &= \left[ \mathbf{k}_n^{(l)} \Vert \mathbf{K}_{\le i}^{(l)} \right] \\
    \tilde{\mathbf{V}}^{(l)}_{\le i} &= \left[ \mathbf{v}_n^{(l)} \Vert \mathbf{V}_{\le i}^{(l)} \right]
\end{split}
\end{equation}
where $\Vert$ denotes concatenation along the sequence dimension. This operation is applied independently to each attention head. The augmented matrices allow every query $\mathbf{q}_i$ to attend to the global summary at virtual position 0, enabling early tokens to access sequence-level context within a single forward pass. Both components are essential: the key enables content-based addressing, while the value carries the aggregated semantic information. We empirically verify that the final token's KV states capture sequence-level semantics through probing experiments in Appendix~\ref{app:probing}. The complete algorithm is provided in Appendix~\ref{app:algorithm}. We further introduce an attention bias $b$ added to the pre-softmax logits when attending to the re-routed position, controlling how strongly each token attends to the global summary. We set $b=1.0$ based on ablation studies.  Detailed formulation is provided in Appendix ~\ref{sec:attention_bias_detail}. 

\subsubsection{Automated Layer Selection via Intrinsic Dimensionality}

The functional specialization of layers necessitates careful selection of the re-routing set $\mathcal{L}$. Shallow layers encode surface-level patterns while final layers are biased toward vocabulary prediction~\citep{skean2025layer}. To identify layers with high semantic density, we leverage Intrinsic Dimensionality (ID), following the observation that the ID minimum corresponds to maximal semantic abstraction~\citep{valeriani2023geometry}. We observe that ID trajectories vary by architecture (see Appendix~\ref{app:id_analysis}), necessitating an adaptive selection strategy.

We employ the TwoNN estimator~\citep{FaccoElena2017Etid} to compute ID across layers using 1,000 sentences from F2LLM~\citep{zhang2025f2llm}. For models exhibiting a clear U-shaped trajectory, we identify the layer $l^*$ achieving minimum ID and define:
\begin{equation}
  \mathcal{L} = \{l \mid l^* \le l \le \min(L, l^* + \lfloor 0.1L \rfloor) \}
\end{equation}
This yields a contiguous window starting from the compression peak. For models with multiple local minima, we instead select layers from low-ID regions in the middle-to-late portion of the network, excluding the first $\lfloor 0.2L \rfloor$ layers that primarily encode surface features. This strategy targets layers where representations are most compressed while avoiding early-layer noise and late-layer prediction bias. We provide empirical validation in Appendix~\ref{app:layer_analysis}.

Finally, the sequence embedding $\mathbf{e}$ is obtained by averaging mean pooling and last-token pooling over the final-layer hidden states, combining distributional evidence across positions with the globally-informed final state, followed by $\ell_2$ normalization.

\section{Experiments}
In this section, we evaluate KV-Embedding across multiple benchmarks and model architectures to demonstrate its efficacy as a training-free framework for generating high-quality text representations.

\begin{table*}[t]
\centering
\small
\resizebox{.93\textwidth}{!}{%
\begin{tabular}{lcccccccc}
\toprule
\textbf{Method} & \textbf{STS} & \textbf{Retr.} & \textbf{Class.} & \textbf{Pair.} & \textbf{Clust.} & \textbf{Rerank.} & \textbf{Summ.} & \textbf{Avg.} \\
\midrule
\multicolumn{9}{c}{\textbf{\textit{Qwen3-4B}}} \\
\midrule
Last Token & 0.2818 & 0.0722 & 0.4243 & 0.3054 & 0.2395 & 0.3476 & 0.2341 & 0.2721 \\
Mean Pooling & 0.4571 & 0.0294 & 0.4656 & 0.5269 & 0.3305 & 0.3272 & 0.1949 & 0.3331 \\
PromptEOL & 0.6741 & 0.1857 & 0.6138 & 0.5637 & 0.3651 & 0.4924 & 0.2396 & 0.4478 \\
Echo & 0.6634 & 0.1727 & 0.5761 & 0.6367 & 0.3530 & 0.4315 & 0.2294 & 0.4375 \\
Token Prepending & 0.6370 & 0.1622 & 0.6283 & 0.5206 & 0.3372 & 0.4901 & 0.2483 & 0.4320 \\


\midrule
KV-Embedding & \textbf{0.7141} & \textbf{0.2765} & \textbf{0.6375} & \textbf{0.6800} & \textbf{0.3903} & \textbf{0.5007} & \textbf{0.2566} & \textbf{0.4937} \\
\quad \textit{w/o KV Re-routing} & 0.5631 & 0.2023 & 0.5094 & 0.5296 & 0.3769 & 0.4600 & 0.2257 & 0.4096\\
\midrule
\multicolumn{9}{c}{\textbf{\textit{Mistral-7B-Instruct-v0.1}}} \\
\midrule
Last Token & 0.3915 & 0.1036 & 0.5314 & 0.3440 & 0.2516 & 0.3673 & 0.2823 & 0.3245 \\
Mean Pooling & 0.4885 & 0.0886 & 0.5157 & 0.5681 & 0.3472 & 0.3790 & 0.2557 & 0.3775 \\
PromptEOL & 0.6953 & 0.1746 & 0.6571 & 0.5629 & 0.3158 & 0.4835 & 0.2863 & 0.4537 \\
Echo & 0.7333 & 0.2414 & 0.6398 & 0.7560 & 0.3681 & 0.4751 & 0.2922 & 0.5008 \\
Token Prepending & 0.6775 & 0.1619 & 0.6580 & 0.5764 & 0.2825 & 0.4790 & 0.2921 & 0.4468 \\


\midrule
KV-Embedding & \textbf{0.7720} & \textbf{0.3014} & \textbf{0.6951} & \textbf{0.7564} & \textbf{0.3902} & \textbf{0.5145} & \textbf{0.3088} & \textbf{0.5341} \\
\quad \textit{w/o KV Re-routing} & 0.6478 & 0.2014 & 0.6008 & 0.5931 & 0.3611 & 0.4804 & 0.2664& 0.4502 \\

\midrule
\multicolumn{9}{c}{\textbf{\textit{Llama-3.1-8B-Instruct}}} \\
\midrule
Last Token & 0.3513 & 0.0887 & 0.4769 & 0.3554 & 0.2652 & 0.3608 & 0.2526 & 0.3073 \\
Mean Pooling & 0.4702 & 0.0713 & 0.4848 & 0.5176 & 0.3577 & 0.3537 & 0.1996 & 0.3507 \\
PromptEOL & 0.6919 & 0.2017 & 0.6250 & 0.5911 & 0.4035 & 0.4991 & 0.2788 & 0.4702 \\
Echo & 0.7124 & 0.2941 & 0.6261 & 0.7125 & 0.4189 & 0.5074 & 0.2527 & 0.5034 \\
Token Prepending & 0.6968 & 0.1817 & 0.6577 & 0.5734 & 0.3365 & 0.4992 & \textbf{0.2880} & 0.4619 \\


\midrule
KV-Embedding & \textbf{0.7398} & \textbf{0.3079} & \textbf{0.6602} & \textbf{0.7332} & \textbf{0.4448} & \textbf{0.5237} & 0.2796 & \textbf{0.5270} \\
\quad \textit{w/o KV Re-routing} &0.6187 &0.2088 & 0.5611 & 0.5839 & 0.4076 & 0.5044 & 0.2746 & 0.4513\\
\bottomrule
\end{tabular}%
}
\caption{Results on MTEB. Each data reports the main metric for the corresponding task category, averaged across datasets. KV-Embedding achieves the best overall performance across all backbones. \textit{w/o KV Re-routing} ablates re-routing while retaining the compression prompt. The highest average score for each model is highlighted in \textbf{bold}. }
\label{tab:embedding_results}
\end{table*}

\subsection{Experimental Setup}

\paragraph{Benchmarks.} We evaluate KV-Embedding on two complementary benchmarks. MTEB~\citep{muennighoff2023mteb} provides a multi-task assessment across seven categories: STS, Retrieval, Classification, Pair Classification, Clustering, Reranking, and Summarization. To evaluate robustness in long-context scenarios, we use LoCoV1~\citep{saad2024locov1}, truncating documents to 1024, 2048, and 4096 tokens. This provides a stress test for KV re-routing, as causal attention typically suffers from context dilution in long sequences. Detailed task descriptions and dataset statistics for these benchmarks are provided in Appendix~\ref{app:benchmarks}.

\paragraph{Models.} To demonstrate the model-agnostic nature of our approach, we experiment with three widely-used decoder-only LLMs covering different parameter scales and families: Qwen3-4B~\citep{yang2025qwen3}, Mistral-7B-Instruct-v0.1~\citep{jiang2023mistral}, and Llama-3.1-8B-Instruct~\citep{dubey2024llama}. All evaluations are performed in a zero-shot, training-free setting.

\paragraph{Baselines.} We compare KV-Embedding against representative training-free strategies. Last Token uses the final hidden state $\mathbf{h}_n$, while Mean Pooling averages all hidden states. PromptEOL~\citep{jiang2024prompteol} guides semantic compression via a template prompt. Echo~\citep{springer2025echo} repeats the input sequence to simulate bidirectionality, doubling the sequence length. Token Prepending~\citep{fu2025tokenprepending} propagates context via special tokens, starting from layer 8 and extracting representations from the 6th-to-final layer. We also report an ablation (\textit{w/o KV Re-routing}) that uses our prompt and pooling strategy but disables the key-value re-routing component. All methods extract representations from the final layer unless otherwise specified. 


\paragraph{Implementation.} We estimate intrinsic dimensionality via TwoNN~\citep{FaccoElena2017Etid} on 1,000 sentences from F2LLM~\citep{zhang2025f2llm}. For Qwen3-4B and Mistral-7B-Instruct-v0.1, which exhibit clear ID minima in middle layers, we apply KV re-routing to contiguous regions: layers 12--21 for Qwen3-4B and layers 13--19 for Mistral-7B-Instruct-v0.1. Llama-3.1-8B shows a more complex trajectory with multiple local minima; we select layers from low-ID regions excluding the shallow minimum (layers 10--11, 20, 26--31). The maximum sequence length is 512 for MTEB. Experiments are conducted on 4 NVIDIA H800 GPUs with batch size 64 and random seed 42.

\subsection{Results}

Table~\ref{tab:embedding_results} and Table~\ref{tab:sequence_length_results} present average results on MTEB and LoCoV1 respectively. The detailed data are provided in the Appendix~\ref{sec:mteb_tables} and~\ref{sec:locov1_results}.

\paragraph{MTEB Results.} As shown in Table~\ref{tab:embedding_results}, KV-Embedding consistently outperforms all baselines across the three backbones. Compared to PromptEOL and Echo, our method achieves higher performance while maintaining the original sequence length. To disentangle the contributions of each component, we ablate KV re-routing while retaining the compression-oriented prompt. The ablation (w/o KV Re-routing) results in a performance drop across all models: on Mistral-7B, the average score decreases from 0.534 to 0.450. This confirms that while the prompt mitigates prediction bias, KV re-routing is the primary driver of improvement.

\paragraph{Impact Across Task Categories.} Performance improvements vary by task type. Tasks requiring holistic semantic understanding show significant gains; notably, Retrieval exhibits the largest increase, reaching 0.2765 on Qwen3-4B compared to 0.1857 for PromptEOL. This suggests that document-level matching benefits from access to global information across the entire sequence. The ablation confirms this pattern: removing re-routing also causes the largest drop in Retrieval. STS and Clustering also show consistent improvements, while Summarization tasks show the smallest variance across methods, with all approaches performing within a similar range.

\begin{table}[ht]
  \centering
  \resizebox{.95\linewidth}{!}{%
  \begin{tabular}{lcccc}
  \toprule
  \textbf{Method} & \textbf{1024} & \textbf{2048} & \textbf{4096} \\
  \midrule
  \multicolumn{4}{l}{\textbf{\textit{Qwen3-4B}}} \\
  \midrule
  PromptEOL & 0.0413 & 0.0438 & 0.1286 \\
  Echo & 0.0961 & 0.0658 & 0.1318 \\
  Token Prepending & 0.0414 & 0.0422 & 0.1255 \\
  KV-Embedding & \textbf{0.1289} & \textbf{0.1192} & \textbf{0.1822} \\
  \midrule
  \multicolumn{4}{l}{\textbf{\textit{Mistral-7B-Instruct-v0.1}}} \\
  \midrule
  PromptEOL& 0.0414 & 0.0407 & 0.0442 \\
  Echo& 0.0627 & 0.0549 & 0.0591 \\
  Token Prepending & 0.0416 & 0.0422 & 0.0992 \\
  KV-Embedding & \textbf{0.2165} & \textbf{0.1838} & \textbf{0.2068} \\
  \midrule
  \multicolumn{4}{l}{\textbf{\textit{Llama-3.1-8B-Instruct}}} \\
  \midrule
  PromptEOL & 0.0463 & 0.0410 & 0.1285 \\
  Echo& 0.1128 & 0.0385 & 0.1191 \\
  Token Prepending & 0.0421 & 0.0422 & 0.1397 \\
  KV-Embedding & \textbf{0.2191} & \textbf{0.1817} & \textbf{0.2404} \\
  \bottomrule
  \end{tabular}%
  }
  \caption{Retrieval performance (NDCG@10) on LoCoV1 across different context lengths. KV-Embedding consistently outperforms all baselines regardless of sequence length.}\label{tab:sequence_length_results}
  \end{table}

\paragraph{Long-Context Evaluation.} Table~\ref{tab:sequence_length_results} presents results on LoCoV1 across different sequence lengths. KV-Embedding substantially outperforms all baselines at every length tested. On Mistral-7B, our method achieves scores above 0.18 across all settings, while baselines remain below 0.10 regardless of sequence length. Similar patterns hold for Qwen3-4B and Llama-3.1-8B, where KV-Embedding consistently outperforms the best baseline by a factor of 1.3--3.5$\times$. These results suggest that baseline methods struggle with long-context retrieval in general, likely because the final token cannot effectively aggregate information from distant positions under causal attention. By explicitly re-routing a global summary to all positions, KV-Embedding enables more effective long-range information integration.

\section{Analysis and Ablation}
In this section, we conduct a series of experiments to explore the KV-Embedding framework, validate its core mechanisms, and analyze the resulting embedding space quality.

\begin{table*}[t]
  \centering
  \small
  \begin{tabular}{p{2.5cm}p{12.5cm}}
  \toprule
  \textbf{Method} & \textbf{Key Tokens with Attention Weights} \\
  \midrule
  \textbf{Echo} & 
  \att{att2}{Looking} \att{att2}{for} \att{att1}{comprehensive} \att{att1}{machine} \att{att0}{learning} \att{att0}{tutorials} \att{att0}{that} \att{att0}{cover} \att{att0}{neural} \att{att0}{networks} \att{att0}{decision} \att{att0}{trees} \att{att0}{and} \att{att0}{deep} \att{att0}{learning} \att{att0}{algorithms} \att{att0}{with} \att{att0}{practical} \att{att0}{Python} \att{att0}{code} \att{att0}{examples} \att{att0}{suitable} \att{att0}{for} \att{att5}{beginners} \\
  \midrule
  \textbf{PromptEOL} & 
  \att{att0}{Looking} \att{att0}{for} \att{att2}{comprehensive} \att{att4}{machine} \att{att2}{learning} \att{att4}{tutorials} \att{att0}{that} \att{att0}{cover} \att{att2}{neural} \att{att0}{networks} \att{att0}{decision} \att{att0}{trees} \att{att0}{and} \att{att0}{deep} \att{att0}{learning} \att{att0}{algorithms} \att{att0}{with} \att{att0}{practical} \att{att0}{Python} \att{att0}{code} \att{att0}{examples} \att{att0}{suitable} \att{att0}{for} \att{att0}{beginners} \\
  \midrule
  \textbf{Token Prepending} & 
  \att{att0}{Looking} \att{att0}{for} \att{att3}{comprehensive} \att{att4}{machine} \att{att3}{learning} \att{att3}{tutorials} \att{att0}{that} \att{att0}{cover} \att{att2}{neural} \att{att0}{networks} \att{att0}{decision} \att{att0}{trees} \att{att0}{and} \att{att0}{deep} \att{att0}{learning} \att{att0}{algorithms} \att{att0}{with} \att{att0}{practical} \att{att0}{Python} \att{att0}{code} \att{att0}{examples} \att{att0}{suitable} \att{att0}{for} \att{att0}{beginners} \\
  \midrule
  \textbf{KV-Embedding} & 
  \att{att6}{Looking} \att{att2}{for} \att{att2}{comprehensive} \att{att1}{machine} \att{att1}{learning} \att{att3}{tutorials} \att{att2}{that} \att{att0}{cover} \att{att0}{neural} \att{att0}{networks} \att{att0}{decision} \att{att0}{trees} \att{att0}{and} \att{att0}{deep} \att{att1}{learning} \att{att0}{algorithms} \att{att0}{with} \att{att0}{practical} \att{att0}{Python} \att{att0}{code} \att{att2}{examples} \att{att2}{suitable} \att{att0}{for} \att{att2}{beginners} \\
  \bottomrule
  \end{tabular}
  \caption{Attention distribution for search query: ``Looking for comprehensive machine learning tutorials that cover neural networks, decision trees, and deep learning algorithms with practical Python code examples suitable for beginners.'' Colors represent attention weights (white: $<$0.012, light: 0.012-0.055, medium: 0.055-0.15, dark: $>$0.15). Key content words are shown.}\label{tab:attention_search_query}
  \end{table*}

\subsection{Mechanism Validation}

\paragraph{Risks of Naive Mask Removal}
A natural question is whether simply removing the causal mask can achieve global context access. As shown in Table~\ref{tab:bi_attn_main}, this approach fails: applying bidirectional attention consistently degrades performance, scoring even lower than the simple last-token baseline on both Qwen3-4B and Mistral-7B-Instruct-v0.1. This failure occurs because decoder-only LLMs are optimized for unidirectional patterns during pre-training; forcing tokens to attend to future positions introduces out-of-distribution states that the model cannot process effectively. In contrast, KV-Embedding relocates existing KV states rather than altering the attention structure, preserving the model's internal manifold while resolving information asymmetry. Further analysis is provided in Appendix~\ref{sec:attention_ablation}.

\begin{table*}[htbp]
  \centering
  \small
  \setlength{\tabcolsep}{0pt}
  \begin{tabular*}{.99\textwidth}{@{\extracolsep{\fill}} ll ccccccccc}
  \toprule
  \textbf{Model} & \textbf{Method} & \textbf{STS} & \textbf{Retr.} & \textbf{Clas.} & \textbf{Pair.} & \textbf{Clust.} & \textbf{Rerank.} & \textbf{Summ.} & \textbf{Avg.} \\
  \midrule
  \multirow{3}{*}{Qwen3-4B} 
   & Last Token + Bi-Attn & 0.1794 & 0.0248 & 0.3764 & 0.2214 & 0.1901 & 0.3146 & 0.2530 & 0.2228 \\
   & Last Token + Causal Attn & 0.2818 & 0.0722 & 0.4243 & 0.3054 & 0.2395 & 0.3476 & 0.2341 & 0.2721 \\
   & KV-Embedding & \textbf{0.7141} & \textbf{0.2765} & \textbf{0.6375} & \textbf{0.6800} & \textbf{0.3903} & \textbf{0.5007} & \textbf{0.2566} & \textbf{0.4937} \\
  \midrule
  \multirow{3}{*}{Mistral-7B-Instruct} 
   & Last Token + Bi-Attn & 0.2255 & 0.0216 & 0.4133 & 0.2748 & 0.2308 & 0.3458 & 0.2746 & 0.2552 \\
   & Last Token + Causal Attn & 0.3915 & 0.1036 & 0.5314 & 0.3440 & 0.2516 & 0.3673 & 0.2823 & 0.3245 \\
   & KV-Embedding & \textbf{0.7720} & \textbf{0.3014} & \textbf{0.6951} & \textbf{0.7564} & \textbf{0.3902} & \textbf{0.5145} & \textbf{0.3088} & \textbf{0.5341} \\
  \bottomrule
  \end{tabular*}
  \caption{Impact of attention constraints. Removing the causal mask (Bi-Attn) causes performance collapse, while KV-Embedding enables global context access without violating causal priors.}\label{tab:bi_attn_main}
  \end{table*}

\begin{figure}[t]
  \centering
  \includegraphics[width=.95\linewidth]{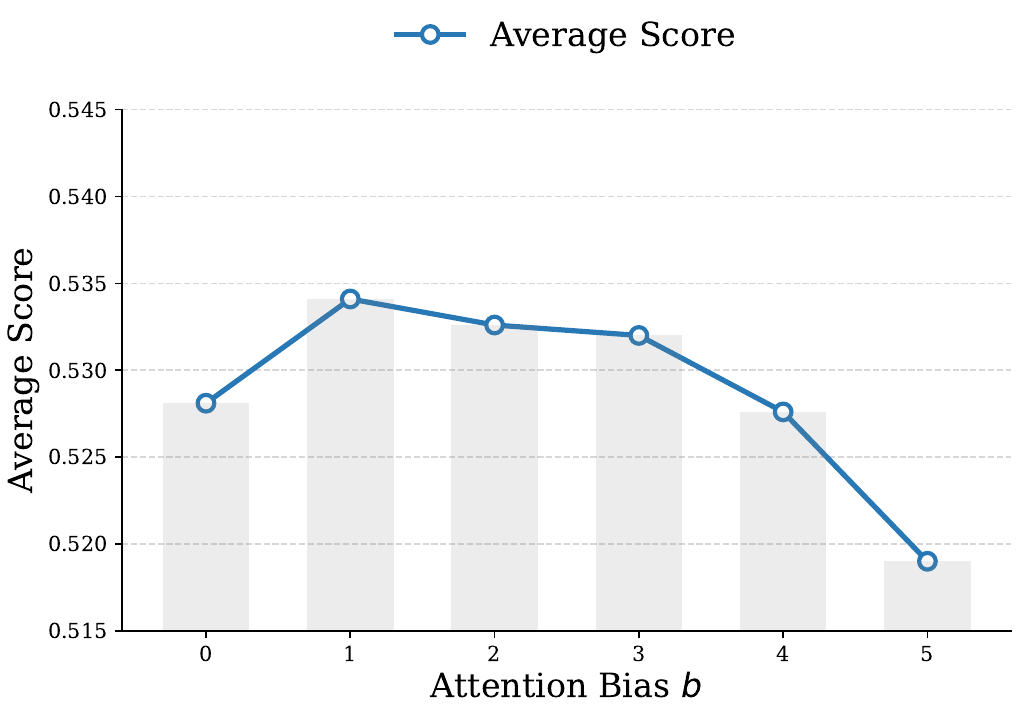}
  \caption{Average performance of Mistral-7B-Instruct with KV-Embedding across different attention bias values. A bias value of 1.0 corresponds to the highest average score.}\label{fig:attn_bias}
\end{figure}

\subsection{Embedding Analysis}



\paragraph{Attention Pattern Visualization}
Table~\ref{tab:attention_search_query} compares the attention distribution of the final token across different methods. Echo exhibits a strong recency bias, focusing primarily on the last tokens. PromptEOL and Token Prepending capture specific keywords but often miss initial intent-bearing words like ``Looking''. In contrast, KV-Embedding demonstrates a more structured alignment with the query; it assigns higher weights to both the intent word ``Looking'' and functional anchors such as ``examples'' and ``beginners''. This pattern indicates that re-routing global context enables the final state to aggregate key semantic components from disparate positions, overcoming the limitations of causal attention.

\paragraph{Embedding Space Quality}

Table~\ref{tab:mistral_alignment_uniformity} presents the alignment and uniformity metrics~\citep{wang2020understanding} for Mistral-7B-Instruct. Alignment evaluates the proximity of embeddings for similar sentence pairs, while uniformity assesses the feature distribution on the hypersphere. KV-Embedding achieves the best values for both metrics compared to the baselines. The improvement in uniformity indicates that internal KV re-routing helps mitigate the anisotropy problem inherent in decoder-only LLMs, where representations tend to cluster within a narrow cone. By redistributing global context, KV-Embedding encourages a more discriminative and isotropic embedding space. Detailed experimental settings regarding the geometric analysis are provided in Appendix~\ref{sec:geometry}.

\begin{table}[ht]
\centering
\resizebox{.99\linewidth}{!}{
  \begin{tabular}{lcc}
    \toprule
    \textbf{Method} & \textbf{Alignment} ($\downarrow$) & \textbf{Uniformity} ($\downarrow$) \\
    \midrule
    PromptEOL & 0.7378 & -2.2715 \\
    Echo & 0.6366 & -2.0794 \\
    Token Prepending & 0.7505 & -2.2233 \\
    KV-Embedding & 0.6082 & -2.3899 \\
    \bottomrule
    \end{tabular}
}
\caption{Alignment and uniformity analysis on Mistral-7B-Instruct-v0.1 with different methods. $\downarrow$ indicates lower is better.}\label{tab:mistral_alignment_uniformity}
\end{table}

\subsection{Ablation of Key Components}
\paragraph{Effect of Attention Bias} 
We add an attention bias $b$ to the pre-softmax logits when attending to the re-routed position, controlling how strongly each token attends to the global summary. As shown in Figure~\ref{fig:attn_bias}, performance improves from $b=0$ to $b=1.0$, confirming that the re-routed context directly enhances embedding quality. The curve remains stable across a wide range, suggesting that the injected KV pairs integrate naturally with the pre-trained attention mechanism. However, performance declines when $b > 3.0$, indicating over-reliance on the global summary. We set $b=1.0$ to balance global context access with local specificity. Detailed formulation and full results are provided in Appendix~\ref{sec:attention_bias_detail}.

\paragraph{Effectiveness of ID-based Layer Selection}
We compare our ID-based strategy with uniform selection targeting the early, middle, or late thirds of the network. As detailed in Appendix~\ref{sec:layer_ablation}, re-routing at early layers yields the lowest performance (0.409 on Qwen3-4B), confirming that shallow representations lack the semantic density required for effective context redistribution. The ID-based strategy achieves the highest scores (0.494 on Qwen3-4B, 0.534 on Mistral-7B) while using fewer layers than uniform alternatives. On Mistral-7B, ID selection (7 layers) outperforms the 10-layer middle-third strategy, demonstrating that the ID minimum effectively identifies layers with peak semantic abstraction.

\paragraph{Pooling Strategy}
We evaluate the impact of different aggregation methods, including last-token, mean, and hybrid pooling. As shown in Appendix~\ref{sec:pooling_ablation}, the hybrid strategy, which averages the representations from last-token and mean pooling, shows greater consistency across different models and tasks. Mean pooling tends to underperform, as it weights all tokens equally without prioritizing the globally-informed final state. Based on these findings, we adopt the hybrid approach in the main experiments.

\paragraph{Robustness to Prompt Design}
To evaluate sensitivity to prompt design, we test five templates varying in prefix and verb choice. Performance remains within a narrow range (0.529--0.540), as shown in Appendix~\ref{sec:prompt_ablation}. This stability suggests that KV-Embedding is relatively insensitive to prompt variations, and that internal KV re-routing may reduce the reliance on specific instructions to elicit semantic representations.

\section{Conclusion}
We introduced KV-Embedding, a training-free framework that extracts text representations from decoder-only LLMs by redistributing their internal states. By re-routing the final token's KV states as a prefix within selected layers, our approach enables all positions to access a compressed sequence summary within a single forward pass, effectively addressing the information asymmetry inherent in causal attention. An automated layer selection strategy based on intrinsic dimensionality enables model-agnostic applicability without manual tuning. Evaluations on MTEB and the long-context retrieval benchmark LoCoV1 demonstrate that KV-Embedding outperforms existing training-free methods by up to 10\%, while maintaining robust performance on sequences up to 4,096 tokens. These results suggest that decoder-only LLMs contain latent representational capacity that can be unlocked through internal state manipulation rather than external input modification. We hope this work encourages further exploration of internal mechanisms in LLMs, bridging the gap between generative pretraining and representation learning.

\section*{Limitations}
Despite the consistent improvements of KV-Embedding over training-free baselines, two main limitations should be noted. First, the re-routing mechanism increases latency compared to standard pooling. Second, as a training-free method, KV-Embedding may not match the performance of supervised models fine-tuned with contrastive objectives on large datasets. Therefore, it can serve as a practical alternative for resource-constrained scenarios rather than a full replacement for state-of-the-art supervised text embeddings.

\bibliography{custom}

\appendix

\section{Algorithm Details}\label{app:algorithm}

Algorithm~\ref{alg:kv_embedding} summarizes the complete KV-Embedding procedure. The method requires only a single forward pass, with KV re-routing applied at the selected layers $\mathcal{L}$.

\begin{algorithm}[htbp]
  \caption{KV-Embedding}
  \label{alg:kv_embedding}
  \small
  \begin{algorithmic}[1]
  \Require Text $x$, LLM with $L$ layers, layer set $\mathcal{L}$
  \Ensure Embedding $\mathbf{e}$
  \State Apply compression prompt to $x$
  \For{$l = 1, \dots, L$}
      \State $\mathbf{K}^{(l)}, \mathbf{V}^{(l)} \gets$ key-value projections
      \If{$l \in \mathcal{L}$}
          \State $\mathbf{k}_n, \mathbf{v}_n \gets$ KV at final position
          \State $\tilde{\mathbf{K}}^{(l)} \gets [\mathbf{k}_n \,\Vert\, \mathbf{K}^{(l)}]$
          \State $\tilde{\mathbf{V}}^{(l)} \gets [\mathbf{v}_n \,\Vert\, \mathbf{V}^{(l)}]$
          \State Attention$(\mathbf{Q}^{(l)}, \tilde{\mathbf{K}}^{(l)}, \tilde{\mathbf{V}}^{(l)})$
      \Else
          \State Standard attention
      \EndIf
  \EndFor
  \State $\mathbf{e}_1 \gets \mathbf{h}_n^{(L)}$, \quad $\mathbf{e}_2 \gets \mathrm{Mean}(\mathbf{H}^{(L)})$
  \State \Return $\mathrm{Normalize}((\mathbf{e}_1 + \mathbf{e}_2)/2)$
  \end{algorithmic}
  \end{algorithm}

\section{Probing KV States for Sequence Information}
\label{app:probing}

To empirically verify that the final token's KV states capture richer sequence-level semantics than earlier positions, we conduct a probing experiment on text classification tasks.

\paragraph{Setup}

We select two classification datasets from MTEB~\citep{muennighoff2023mteb}: ImdbClassification (binary sentiment) and TweetSentimentExtractionClassification (three-way sentiment). For each dataset, we sample 1,000 training instances and 500 validation instances. We perform a forward pass through Mistral-7B-Instruct~\citep{jiang2023mistral} and extract KV states from three positions: the first token, the middle token, and the last token. For each position, we concatenate the key and value vectors across all attention heads at the final layer, yielding a fixed-dimensional representation. A logistic regression classifier is trained on these frozen representations, with the L2 regularization coefficient selected via cross-validation over the range $[10^{-4}, 10^1]$.

\paragraph{Results}

Table~\ref{tab:probing} presents the probing accuracy for each position. Across both datasets, the last token consistently achieves the highest accuracy. On IMDB, the accuracy improves from 51.4\% at the first position to 85.2\% at the last position, a gain of over 33 absolute points. A similar trend is observed on Tweet sentiment classification, where the last position outperforms the first by 15.6 points. These results confirm that sequence-level information progressively accumulates toward the end under causal attention, providing direct empirical support for our KV re-routing approach.

\begin{table}[h]
\centering
\small
\resizebox{0.8\linewidth}{!}{%
\begin{tabular}{lcc}
\toprule
\textbf{Position} & \textbf{IMDB} & \textbf{TweetSentiment} \\
\midrule
First & 51.4 & 45.2 \\
Middle & 73.0 & 52.6 \\
Last & 85.2 & 60.8 \\
\bottomrule
\end{tabular}
}
\caption{Probing accuracy (\%) on classification tasks using KV states from different token positions in Mistral-7B-Instruct.}
\label{tab:probing}
\end{table}

\section{Layer-wise Representation Analysis}
\label{app:layer_analysis}

To understand how representation quality evolves across layers and validate our layer selection strategy, we evaluate the last token's hidden states at each layer on semantic similarity tasks using Mistral-7B-Instruct~\citep{jiang2023mistral}.

\paragraph{Setup}

We extract hidden states from all 32 layers and evaluate on two tasks: Pair Classification, which measures fine-grained semantic matching via cosine AP, and STS, which measures similarity via Spearman correlation. These tasks reveal complementary aspects of representation quality.

\paragraph{Results}

Table~\ref{tab:layer_analysis} reveals distinct patterns for the two tasks. Pair Classification peaks at layers 10--13 and declines thereafter, indicating that middle layers capture the most discriminative features for fine-grained semantic matching. In contrast, STS improves toward later layers, likely due to the next-token prediction bias that emphasizes local word associations over holistic semantics.

\begin{table}[h]
\centering
\resizebox{0.85\linewidth}{!}{%
\begin{tabular}{lcc}
\toprule
\textbf{Layer Range} & \textbf{Pair Class.} & \textbf{STS} \\
\midrule
0--6 & 0.270 & 0.221 \\
7--12 & 0.302 & 0.282 \\
13--19 (ID-selected) & 0.301 & 0.290 \\
20--26 & 0.286 & 0.297 \\
27--32 & 0.275 & 0.339 \\
\bottomrule
\end{tabular}%
}
\caption{Layer-wise probing performance on Mistral-7B-Instruct. The ID-selected range maintains strong average performance while avoiding the generation-biased later layers.}
\label{tab:layer_analysis}
\end{table}

Our ID-based strategy selects layers 13--19, corresponding to a transition zone where semantic abstraction is high but prediction bias has not yet dominated. This range avoids early layers with insufficient semantic content and final layers increasingly optimized for next-token prediction. The Pair Classification results support this choice, confirming that the ID-selected range preserves strong semantic matching capability.

\section{Benchmark Details}
\label{app:benchmarks}

\subsection{MTEB Benchmark}
The Massive Text Embedding Benchmark (MTEB)~\citep{muennighoff2023mteb} provides a standardized framework for evaluating text embeddings across diverse semantic scenarios. We conduct a comprehensive evaluation covering seven core task categories, encompassing 42 datasets to ensure a holistic assessment of semantic representations.

\paragraph{Semantic Textual Similarity (STS)} Results are reported on the full suite of standard STS tasks, including BIOSSES, SICK-R, and STS12 through STS16. Following standard practice, we use the Spearman correlation between the cosine similarity of embedding pairs and human-annotated gold labels as the core metric for measuring fine-grained alignment in the embedding space.

\paragraph{Reranking and Retrieval} Reranking performance is measured on AskUbuntuDupQuestions, MindSmallReranking, SciDocsRR, and StackOverflowDupQuestions using Mean Average Precision (MAP). For retrieval, we employ a broad set of benchmarks including NFCorpus, FiQA2018, SciFact, and QuoraRetrieval, where NDCG@10 serves as the primary metric for evaluating ranking effectiveness.

\paragraph{Classification and Summarization} For classification tasks, we evaluate the discriminative capacity of embeddings on ToxicConversations, Imdb, Banking77, Emotion, and TweetSentimentExtraction, reporting the accuracy score for each. The summarization capability is measured via the SummEval dataset, where we calculate the Spearman correlation based on cosine similarity to evaluate the alignment between generated summaries and source documents.

\paragraph{Clustering and Pair Classification} The clustering evaluation encompasses eleven datasets across diverse domains, including Biorxiv, Medrxiv, Reddit, and StackExchange, where performance is quantified using the V-measure. For pair classification tasks, such as TwitterURLCorpus and SprintDuplicateQuestions, we report Average Precision (AP) to measure the sensitivity of the model to binary semantic similarity.

\subsection{LoCoV1 Benchmark}
LoCoV1~\citep{saad2024locov1} serves as a specialized stress test for long-context stability. Unlike the predominantly short-form text in MTEB, LoCoV1 requires maintaining semantic density over extended sequences. To analyze the influence of sequence length on our KV re-routing mechanism, we evaluate performance under controlled truncation at 1k, 2k, and 4k tokens. We adopt NDCG@10 for retrieval tasks, comparing the results against different baselines.

\section{Detailed MTEB Results}\label{sec:mteb_tables}

This section presents the detailed results for each task category in the MTEB benchmark~\citep{muennighoff2023mteb}. We report performance across three models: Qwen3-4B, Mistral-7B-Instruct-v0.1, and Llama-3.1-8B-Instruct. We compare six different embedding methods: Last Token, Mean Pooling, PromptEOL~\citep{jiang2024prompteol}, Echo~\citep{springer2025echo}, Token Prepending~\citep{fu2025tokenprepending}, and our proposed KV-Embedding.

The detailed breakdown of results is organized as follows. Table~\ref{tab:sts_results} reports the Spearman correlation scores for Semantic Textual Similarity (STS) tasks, while Table~\ref{tab:retrieval_results} details the Retrieval performance measured by NDCG@10. Classification accuracy across the evaluated datasets is shown in Table~\ref{tab:classification_results}. For Pair Classification, Table~\ref{tab:pair_classification_results} provides the Average Precision (AP) scores. Additionally, we present Clustering results measured by the Validity Measure (V-measure) in Table~\ref{tab:clustering_results}, Reranking performance (Mean Average Precision) in Table~\ref{tab:reranking_results}, and finally, the Summarization Spearman correlations in Table~\ref{tab:summarization_results}.

\begin{table*}[htbp]
  \centering
  \small
  \resizebox{\textwidth}{!}{
  \begin{tabular}{llccccccccc}
  \toprule
  \textbf{Model} & \textbf{Method} & \textbf{Avg.} & \textbf{BIOSSES} & \textbf{SICK-R} & \textbf{STS12} & \textbf{STS13} & \textbf{STS14} & \textbf{STS15} & \textbf{STS16} & \textbf{STS-B} \\
  \midrule
  \multirow{6}{*}{Qwen3-4B} 
   & Last Token & 0.2818 & 0.2781 & 0.4939 & 0.1761 & 0.2796 & 0.1560 & 0.1161 & 0.4477 & 0.3068 \\
   & Mean Pooling & 0.4571 & 0.5943 & 0.5148 & 0.4526 & 0.3556 & 0.3286 & 0.5481 & 0.5249 & 0.3378 \\
   & PromptEOL & 0.6741 & 0.5997 & 0.6612 & 0.5428 & 0.7460 & 0.6356 & 0.7359 & 0.7541 & 0.7176 \\
   & Echo & 0.6634 & 0.6726 & 0.6565 & 0.5796 & 0.6884 & 0.5972 & 0.7275 & 0.7246 & 0.6607 \\
   & Token Prepending & 0.6370 & 0.5648 & 0.6405 & 0.5154 & 0.7040 & 0.6061 & 0.6947 & 0.6924 & 0.6785 \\
   & KV-Embedding & \textbf{0.7141} & 0.7551 & 0.6835 & 0.6247 & 0.7740 & 0.6665 & 0.7622 & 0.7586 & 0.6887 \\
  \midrule
  \multirow{6}{*}{Mistral-7B-Instruct-v0.1} 
   & Last Token & 0.3915 & 0.3089 & 0.5465 & 0.2552 & 0.3657 & 0.2411 & 0.2488 & 0.6033 & 0.5629 \\
   & Mean Pooling & 0.4885 & 0.6425 & 0.5036 & 0.3910 & 0.4536 & 0.3781 & 0.5854 & 0.5695 & 0.3843 \\
   & PromptEOL & 0.6953 & 0.5817 & 0.6918 & 0.6554 & 0.7783 & 0.6740 & 0.7575 & 0.6981 & 0.7258 \\
   & Echo & 0.7333 & 0.7716 & 0.7245 & 0.6041 & 0.7916 & 0.6822 & 0.7899 & 0.7581 & 0.7443 \\
   & Token Prepending & 0.6775 & 0.5716 & 0.6825 & 0.6210 & 0.7587 & 0.6633 & 0.7343 & 0.6741 & 0.7145 \\
   & KV-Embedding & \textbf{0.7720} & 0.7496 & 0.7683 & 0.6891 & 0.8056 & 0.7308 & 0.8254 & 0.7975 & 0.8097 \\
  \midrule
  \multirow{6}{*}{Llama-3.1-8B-Instruct} 
   & Last Token & 0.3513 & 0.3302 & 0.5315 & 0.1837 & 0.3846 & 0.2226 & 0.2052 & 0.5462 & 0.4063 \\
   & Mean Pooling & 0.4702 & 0.6253 & 0.5103 & 0.3583 & 0.4200 & 0.3602 & 0.5650 & 0.5403 & 0.3824 \\
   & PromptEOL & 0.6919 & 0.6509 & 0.7034 & 0.5765 & 0.7376 & 0.6306 & 0.7383 & 0.7636 & 0.7343 \\
   & Echo & 0.7124 & 0.7704 & 0.6956 & 0.5428 & 0.7843 & 0.6514 & 0.7557 & 0.7676 & 0.7311 \\
   & Token Prepending & 0.6968 & 0.6536 & 0.6959 & 0.5947 & 0.7642 & 0.6665 & 0.7416 & 0.7342 & 0.7240 \\
   & KV-Embedding & \textbf{0.7398} & 0.7710 & 0.7254 & 0.6239 & 0.7793 & 0.6946 & 0.7819 & 0.7874 & 0.7546 \\
  \bottomrule
  \end{tabular}
  }
  \caption{MTEB STS Spearman correlation results across different models and methods. The highest average score for each model is highlighted in bold.}
  \label{tab:sts_results}
  \end{table*}

\begin{table*}[htbp]
  \centering
  \small
  \setlength{\tabcolsep}{3.5pt}
  \resizebox{\textwidth}{!}{
  \begin{tabular}{llcccccccccc}
  \toprule
  \textbf{Model} & \textbf{Method} & \textbf{Avg.} & \textbf{ArguAna} & \textbf{NFCorpus} & \textbf{FiQA} & \textbf{SciFact} & \textbf{SCIDOCS} & \textbf{Quora} & \textbf{TREC} & \textbf{Touche} & \textbf{CQADupstack} \\
  \midrule
  \multirow{6}{*}{Qwen3-4B} 
    & Last Token & 0.0722 & 0.0797 & 0.0175 & 0.0000 & 0.0087 & 0.0031 & 0.5140 & 0.0159 & 0.0000 & 0.0111 \\
    & Mean Pooling & 0.0294 & 0.1674 & 0.0142 & 0.0047 & 0.0362 & 0.0034 & 0.0170 & 0.0193 & 0.0000 & 0.0021 \\
    & PromptEOL & 0.1857 & 0.2462 & 0.0948 & 0.0519 & 0.2138 & 0.0850 & 0.6500 & 0.2393 & 0.0135 & 0.0767 \\
    & Echo & 0.1727 & 0.2336 & 0.0347 & 0.0552 & 0.1146 & 0.0211 & 0.7465 & 0.2239 & 0.0327 & 0.0918 \\
    & Token Prepending & 0.1622 & 0.1912 & 0.1060 & 0.0496 & 0.1294 & 0.0713 & 0.5622 & 0.1949 & 0.0066 & 0.1489 \\
    & KV-Embedding & \textbf{0.2765} & 0.4643 & 0.1069 & 0.1023 & 0.4054 & 0.0608 & 0.6916 & 0.3957 & 0.0869 & 0.1749 \\
  \midrule
  \multirow{6}{*}{Mistral-7B-Instruct-v0.1} 
    & Last Token & 0.1036 & 0.0961 & 0.0347 & 0.0141 & 0.0220 & 0.0027 & 0.6655 & 0.0770 & 0.0022 & 0.0178 \\
    & Mean Pooling & 0.0886 & 0.3972 & 0.0201 & 0.0095 & 0.2365 & 0.0044 & 0.0525 & 0.0669 & 0.0000 & 0.0100 \\
    & PromptEOL & 0.1746 & 0.1030 & 0.1404 & 0.0781 & 0.2033 & 0.0934 & 0.5847 & 0.2212 & 0.0359 & 0.1118 \\
    & Echo & 0.2414 & 0.2719 & 0.1199 & 0.0909 & 0.3325 & 0.0327 & 0.7676 & 0.3283 & 0.0487 & 0.1797 \\
    & Token Prepending & 0.1619 & 0.0589 & 0.1453 & 0.0768 & 0.1838 & 0.0834 & 0.5520 & 0.2143 & 0.0173 & 0.1257 \\
    & KV-Embedding & \textbf{0.3014} & 0.2543 & 0.2257 & 0.2063 & 0.3723 & 0.0969 & 0.7774 & 0.4387 & 0.1120 & 0.2292 \\
  \midrule
  \multirow{6}{*}{Llama-3.1-8B-Instruct} 
    & Last Token & 0.0887 & 0.1104 & 0.0210 & 0.0088 & 0.0136 & 0.0037 & 0.6128 & 0.0232 & 0.0007 & 0.0041 \\
    & Mean Pooling & 0.0713 & 0.4113 & 0.0164 & 0.0023 & 0.1201 & 0.0030 & 0.0275 & 0.0593 & 0.0000 & 0.0017 \\
    & PromptEOL & 0.2017 & 0.2983 & 0.1317 & 0.0747 & 0.2140 & 0.0740 & 0.6890 & 0.2316 & 0.0066 & 0.0957 \\
    & Echo & 0.2941 & 0.3685 & 0.1183 & 0.1255 & 0.4756 & 0.0679 & 0.8026 & 0.4382 & 0.0581 & 0.1920 \\
    & Token Prepending & 0.1817 & 0.0721 & 0.1893 & 0.0833 & 0.1678 & 0.1222 & 0.6488 & 0.1952 & 0.0131 & 0.1434 \\
    & KV-Embedding & \textbf{0.3079} & 0.4965 & 0.2046 & 0.1507 & 0.4452 & 0.0724 & 0.7794 & 0.3993 & 0.0602 & 0.1629 \\
  \bottomrule
  \end{tabular}
  }
  \caption{MTEB Retrieval (NDCG@10) results. The highest average score for each model is highlighted in bold. Abbreviations: \textit{TREC} (TREC-COVID), \textit{CQADupstack} (CQADupstackEnglishRetrieval).}
  \label{tab:retrieval_results}
  \end{table*}

\begin{table*}[htbp]
\centering
\small
\setlength{\tabcolsep}{4pt}
\resizebox{\textwidth}{!}{
\begin{tabular}{llcccccc}
\toprule
\textbf{Model} & \textbf{Method} & \textbf{Avg.} & \textbf{Toxic} & \textbf{Imdb} & \textbf{Banking77} & \textbf{Emotion} & \textbf{TweetSent} \\
\midrule
\multirow{6}{*}{Qwen3-4B} 
& Last Token & 0.4243 & 0.6463 & 0.5616 & 0.1850 & 0.2636 & 0.4648 \\
& Mean Pooling & 0.4656 & 0.5867 & 0.6914 & 0.3730 & 0.2244 & 0.4527 \\
& PromptEOL & 0.6138 & 0.6976 & 0.7377 & 0.5057 & 0.5043 & 0.6236 \\
& Echo & 0.5761 & 0.6740 & 0.6315 & 0.6426 & 0.4049 & 0.5278 \\
& Token Prepending & 0.6283 & 0.6837 & 0.8281 & 0.4863 & 0.5079 & 0.6357 \\
& KV-Embedding & \textbf{0.6375} & 0.7088 & 0.7259 & 0.6746 & 0.4859 & 0.5922 \\
\midrule
\multirow{6}{*}{Mistral-7B-Instruct-v0.1} 
& Last Token & 0.5314 & 0.6375 & 0.6324 & 0.5491 & 0.2958 & 0.5424 \\
& Mean Pooling & 0.5157 & 0.6180 & 0.7334 & 0.4702 & 0.2553 & 0.5019 \\
& PromptEOL & 0.6571 & 0.7202 & 0.8641 & 0.5416 & 0.5042 & 0.6554 \\
& Echo & 0.6398 & 0.6868 & 0.7290 & 0.7270 & 0.4452 & 0.6109 \\
& Token Prepending & 0.6580 & 0.7094 & 0.8654 & 0.5572 & 0.5034 & 0.6546 \\
& KV-Embedding & \textbf{0.6951} & 0.7101 & 0.8655 & 0.7204 & 0.5193 & 0.6602 \\
\midrule
\multirow{6}{*}{Llama-3.1-8B-Instruct} 
& Last Token & 0.4769 & 0.6170 & 0.6163 & 0.3641 & 0.2753 & 0.5120 \\
& Mean Pooling & 0.4848 & 0.5940 & 0.7286 & 0.3796 & 0.2414 & 0.4801 \\
& PromptEOL & 0.6250 & 0.6880 & 0.8283 & 0.5530 & 0.4381 & 0.6179 \\
& Echo & 0.6261 & 0.6814 & 0.7328 & 0.6958 & 0.4418 & 0.5788 \\
& Token Prepending & 0.6577 & 0.7065 & 0.8926 & 0.5718 & 0.4796 & 0.6379 \\
& KV-Embedding & \textbf{0.6602} & 0.6931 & 0.8660 & 0.6876 & 0.4605 & 0.5939 \\
\bottomrule
\end{tabular}
}
\caption{MTEB Classification (Accuracy) results. The highest average score for each model is highlighted in bold. Column headers are abbreviated: \textit{Toxic} (ToxicConversations), \textit{TweetSent} (TweetSentimentExtraction).}
\label{tab:classification_results}
\end{table*}

\begin{table*}[htbp]
\centering
\small
\setlength{\tabcolsep}{5pt}
\begin{tabular}{llcccc}
\toprule
\textbf{Model} & \textbf{Method} & \textbf{Avg.} & \textbf{TwitterURL} & \textbf{SprintDup} & \textbf{TwitSemEval} \\
\midrule
\multirow{6}{*}{Qwen3-4B} 
& Last Token & 0.3054 & 0.4003 & 0.1278 & 0.3882 \\
& Mean Pooling & 0.5269 & 0.6720 & 0.5726 & 0.3361 \\
& PromptEOL & 0.5637 & 0.7592 & 0.3597 & 0.5722 \\
& Echo & 0.6367 & 0.6791 & 0.7071 & 0.5239 \\
& Token Prepending & 0.5206 & 0.7637 & 0.2008 & 0.5972 \\
& KV-Embedding & \textbf{0.6800} & 0.7898 & 0.7144 & 0.5357 \\
\midrule
\multirow{6}{*}{Mistral-7B-Instruct-v0.1} 
& Last Token & 0.3440 & 0.4417 & 0.1441 & 0.4462 \\
& Mean Pooling & 0.5681 & 0.7388 & 0.5881 & 0.3774 \\
& PromptEOL & 0.5629 & 0.8079 & 0.1749 & 0.7059 \\
& Echo & 0.7560 & 0.8059 & 0.7916 & 0.6707 \\
& Token Prepending & 0.5764 & 0.8113 & 0.1974 & 0.7206 \\
& KV-Embedding & \textbf{0.7564} & 0.8460 & 0.6939 & 0.7291 \\
\midrule
\multirow{6}{*}{Llama-3.1-8B-Instruct} 
& Last Token & 0.3554 & 0.4745 & 0.1289 & 0.4630 \\
& Mean Pooling & 0.5176 & 0.7157 & 0.4670 & 0.3701 \\
& PromptEOL & 0.5911 & 0.7624 & 0.3968 & 0.6140 \\
& Echo & 0.7125 & 0.7749 & 0.7314 & 0.6312 \\
& Token Prepending & 0.5734 & 0.7842 & 0.2385 & 0.6975 \\
& KV-Embedding & \textbf{0.7332} & 0.8161 & 0.7237 & 0.6598 \\
\bottomrule
\end{tabular}

\caption{MTEB Pair Classification (Average Precision) results. The highest average score for each model is highlighted in bold. Column headers are abbreviated: \textit{SprintDup} (SprintDuplicateQuestions), \textit{TwitSemEval} (TwitterSemEval2015).}
\label{tab:pair_classification_results}
\end{table*}

\begin{table*}[htbp]
  \centering
  \small
  \setlength{\tabcolsep}{3.5pt}
  \resizebox{\textwidth}{!}{
  \begin{tabular}{llcccccccccc}
  \toprule
  \textbf{Model} & \textbf{Method} & \textbf{Avg.} & \textbf{BioP2P} & \textbf{BioS2S} & \textbf{MedP2P} & \textbf{MedS2S} & \textbf{Reddit} & \textbf{RedditP2P} & \textbf{StackEx} & \textbf{StackP2P} & \textbf{20News} \\
  \midrule
  \multirow{6}{*}{Qwen3-4B} 
    & Last Token & 0.2395 & 0.1783 & 0.1737 & 0.2059 & 0.2238 & 0.1610 & 0.3362 & 0.4269 & 0.3053 & 0.1441 \\
    & Mean Pooling & 0.3305 & 0.3808 & 0.2528 & 0.3423 & 0.2806 & 0.1984 & 0.5372 & 0.4074 & 0.4088 & 0.1661 \\
    & PromptEOL & 0.3651 & 0.2855 & 0.3095 & 0.2699 & 0.2989 & 0.3692 & 0.4943 & 0.5488 & 0.3500 & 0.3597 \\
    & Echo & 0.3530 & 0.3290 & 0.2688 & 0.3085 & 0.2909 & 0.3228 & 0.4843 & 0.5631 & 0.3786 & 0.2306 \\
    & Token Prepending & 0.3372 & 0.2851 & 0.3020 & 0.2681 & 0.3009 & 0.2799 & 0.4776 & 0.4002 & 0.3449 & 0.3757 \\
    & KV-Embedding & \textbf{0.3903} & 0.3627 & 0.3165 & 0.2967 & 0.3181 & 0.3721 & 0.5526 & 0.5409 & 0.3890 & 0.3642 \\

  \midrule
  \multirow{6}{*}{Mistral-7B-Instruct-v0.1} 
    & Last Token & 0.2516 & 0.1043 & 0.1821 & 0.1824 & 0.2312 & 0.2666 & 0.3505 & 0.4848 & 0.2860 & 0.1766 \\
    & Mean Pooling & 0.3472 & 0.3981 & 0.2696 & 0.3523 & 0.2889 & 0.2258 & 0.5847 & 0.4004 & 0.4078 & 0.1969 \\
    & PromptEOL & 0.3158 & 0.2289 & 0.2779 & 0.2253 & 0.2671 & 0.3014 & 0.4622 & 0.4519 & 0.3301 & 0.2978 \\
    & Echo & 0.3681 & 0.3407 & 0.2723 & 0.3183 & 0.2804 & 0.3496 & 0.5288 & 0.5566 & 0.3724 & 0.2942 \\
    & Token Prepending & 0.2825 & 0.2124 & 0.2757 & 0.2156 & 0.2580 & 0.2173 & 0.4431 & 0.3192 & 0.3169 & 0.2841 \\
    & KV-Embedding & \textbf{0.3902} & 0.3196 & 0.3308 & 0.2795 & 0.3107 & 0.3901 & 0.5538 & 0.5460 & 0.3685 & 0.4129 \\

  \midrule
  \multirow{6}{*}{Llama-3.1-8B-Instruct} 
    & Last Token & 0.2652 & 0.2125 & 0.1686 & 0.2169 & 0.2275 & 0.2741 & 0.3376 & 0.4912 & 0.2957 & 0.1630 \\
    & Mean Pooling & 0.3577 & 0.4047 & 0.2820 & 0.3516 & 0.2860 & 0.2463 & 0.5870 & 0.4322 & 0.4237 & 0.2060 \\
    & PromptEOL & 0.4035 & 0.3357 & 0.3406 & 0.2916 & 0.3291 & 0.4301 & 0.5354 & 0.5873 & 0.3543 & 0.4269 \\
    & Echo & 0.4189 & 0.3574 & 0.3071 & 0.3303 & 0.3122 & 0.4419 & 0.5893 & 0.6279 & 0.4153 & 0.3891 \\
    & Token Prepending & 0.3365 & 0.3037 & 0.3165 & 0.2672 & 0.3019 & 0.2847 & 0.4788 & 0.3781 & 0.3413 & 0.3559 \\
    & KV-Embedding & \textbf{0.4448} & 0.3770 & 0.3654 & 0.3252 & 0.3453 & 0.4908 & 0.6061 & 0.6103 & 0.3992 & 0.4838 \\
  \bottomrule
  \end{tabular}
  }
\caption{MTEB Clustering (V-measure) results. The highest average score for each model is highlighted in bold. Column headers are abbreviated to fit the page (e.g., \textit{BioP2P} for BiorxivClusteringP2P, \textit{20News} for TwentyNewsgroupsClustering).}
\label{tab:clustering_results}
\end{table*}
  
\begin{table*}[htbp]
\centering
\small
\setlength{\tabcolsep}{6pt} 
\resizebox{\textwidth}{!}{
\begin{tabular}{llccccc}
\toprule
\textbf{Model} & \textbf{Method} & \textbf{Avg.} & \textbf{AskUbuntu} & \textbf{MindSmall} & \textbf{SciDocs} & \textbf{StackOverflow} \\
\midrule
\multirow{6}{*}{Qwen3-4B} 
 & Last Token & 0.3476 & 0.4255 & 0.2445 & 0.4790 & 0.2415 \\
 & Mean Pooling & 0.3272 & 0.4378 & 0.2823 & 0.3543 & 0.2342 \\
 & PromptEOL & 0.4924 & 0.5360 & 0.2947 & 0.7558 & 0.3833 \\
 & Echo & 0.4315 & 0.5065 & 0.2791 & 0.5523 & 0.3880 \\
 & Token Prepending & 0.4901 & 0.5263 & 0.2930 & 0.7573 & 0.3838 \\
 & KV-Embedding & \textbf{0.5007} & 0.5672 & 0.3066 & 0.7203 & 0.4088 \\
\midrule
\multirow{6}{*}{Mistral-7B-Instruct-v0.1} 
 & Last Token & 0.3673 & 0.4432 & 0.2532 & 0.4964 & 0.2763 \\
 & Mean Pooling & 0.3790 & 0.4760 & 0.2715 & 0.4777 & 0.2906 \\
 & PromptEOL & 0.4835 & 0.5402 & 0.2905 & 0.7276 & 0.3759 \\
 & Echo & 0.4751 & 0.5390 & 0.2863 & 0.6402 & 0.4347 \\
 & Token Prepending & 0.4790 & 0.5403 & 0.2776 & 0.7120 & 0.3863 \\
 & KV-Embedding & \textbf{0.5145} & 0.5723 & 0.3034 & 0.7533 & 0.4289 \\
\midrule
\multirow{6}{*}{Llama-3.1-8B-Instruct} 
 & Last Token & 0.3608 & 0.4447 & 0.2458 & 0.4866 & 0.2661 \\
 & Mean Pooling & 0.3537 & 0.4510 & 0.2657 & 0.4254 & 0.2729 \\
 & PromptEOL & 0.4991 & 0.5360 & 0.3095 & 0.7606 & 0.3902 \\
 & Echo & 0.5074 & 0.5338 & 0.3147 & 0.7500 & 0.4313 \\
 & Token Prepending & 0.4992 & 0.5533 & 0.2942 & 0.7458 & 0.4034 \\
 & KV-Embedding & \textbf{0.5237} & 0.5612 & 0.3097 & 0.7841 & 0.4396 \\
\bottomrule
\end{tabular}
}
\caption{MTEB Reranking (Mean Average Precision) results. The highest average score for each model is highlighted in bold. Column headers are abbreviated: \textit{AskUbuntu} (AskUbuntuDupQuestions), \textit{StackOverflow} (StackOverflowDupQuestions), \textit{SciDocs} (SciDocsRR).}
\label{tab:reranking_results}
\end{table*}

\begin{table}[htbp]
\centering
\small
\resizebox{\columnwidth}{!}{
\begin{tabular}{llc}
\toprule
\textbf{Model} & \textbf{Method} & \textbf{SummEval} \\
\midrule
\multirow{6}{*}{Mistral-7B-Instruct-v0.1} 
  & Last Token & 0.2823 \\
  & Mean Pooling & 0.2557 \\
  & PromptEOL & 0.2863 \\
  & Echo & 0.2922 \\
  & Token Prepending & 0.2921 \\
  & KV-Embedding & \textbf{0.3088} \\
\midrule
\multirow{6}{*}{Qwen3-4B} 
  & Last Token & 0.2341 \\
  & Mean Pooling & 0.1949 \\
  & PromptEOL & 0.2396 \\
  & Echo & 0.2294 \\
  & Token Prepending & 0.2483 \\
  & KV-Embedding & \textbf{0.2566} \\
\midrule
\multirow{6}{*}{Llama-3.1-8B-Instruct} 
  & Last Token & 0.2526 \\
  & Mean Pooling & 0.1996 \\
  & PromptEOL & 0.2788 \\
  & Echo & 0.2527 \\
  & Token Prepending & \textbf{0.2880} \\
  & KV-Embedding & 0.2796 \\
\bottomrule
\end{tabular}
}
\caption{MTEB Summarization (Spearman correlation) results. The highest score for each model is highlighted in bold.}\label{tab:summarization_results}
\end{table}

\section{Detailed LoCoV1 Results}\label{sec:locov1_results}
The full results for LoCoV1 experiments are presented in Tables~\ref{tab:loco_1024}, \ref{tab:loco_2048}, and \ref{tab:loco_4096}. Across all three tested context windows and model architectures, our KV-Embedding method consistently achieves the highest average performance, demonstrating better robustness in long-context retrieval tasks.

\begin{table*}[htbp]
\centering
\resizebox{\textwidth}{!}{%
\begin{tabular}{llccccccccccccc}
\toprule
\multirow{2}{*}{\textbf{Model}} & \multirow{2}{*}{\textbf{Method}} & \textbf{2wiki} & \textbf{court} & \textbf{court} & \textbf{gov} & \textbf{legal} & \textbf{multi} & \textbf{pass} & \textbf{qasper} & \textbf{qasper} & \multirow{2}{*}{\textbf{qmsum}} & \textbf{stack} & \textbf{summ} & \multirow{2}{*}{\textbf{Avg.}} \\
& & \textbf{mqa} & \textbf{HTML} & \textbf{Text} & \textbf{rep} & \textbf{case} & \textbf{qa} & \textbf{ret} & \textbf{abs} & \textbf{title} & & \textbf{over} & \textbf{scrn} & \\
\midrule
\multirow{4}{*}{Mistral-7B} 
  & PromptEOL & 0.0757 & 0.0020 & 0.0010 & 0.0059 & 0.0060 & 0.1515 & 0.0757 & 0.0047 & 0.0109 & 0.1492 & 0.0009 & 0.0134 & 0.0414 \\
  & Echo & 0.1699 & 0.0034 & 0.0055 & 0.0047 & 0.0089 & 0.1515 & 0.1461 & 0.0703 & 0.0649 & 0.1033 & 0.0085 & 0.0151 & 0.0627 \\
  & Token Prepending & 0.0757 & 0.0023 & 0.0023 & 0.0047 & 0.0059 & 0.1515 & 0.0757 & 0.0109 & 0.0109 & 0.1492 & 0.0028 & 0.0077 & 0.0416 \\
  & KV-Embedding & 0.1983 & 0.0077 & 0.0090 & 0.2023 & 0.0163 & 0.6080 & 0.1656 & 0.1654 & 0.0338 & 0.3936 & 0.6191 & 0.1791 & 0.2165 \\
\midrule
\multirow{4}{*}{Qwen3-4B} 
  & PromptEOL & 0.0757 & 0.0045 & 0.0025 & 0.0000 & 0.0206 & 0.1515 & 0.0757 & 0.0109 & 0.0109 & 0.1289 & 0.0000 & 0.0144 & 0.0413 \\
  & Echo & 0.3677 & 0.0106 & 0.0078 & 0.0047 & 0.0097 & 0.1515 & 0.0678 & 0.1034 & 0.2250 & 0.1673 & 0.0248 & 0.0134 & 0.0961 \\
  & Token Prepending & 0.0674 & 0.0023 & 0.0023 & 0.0047 & 0.0059 & 0.1515 & 0.0757 & 0.0109 & 0.0109 & 0.1492 & 0.0029 & 0.0134 & 0.0414 \\
  & KV-Embedding & 0.1155 & 0.0052 & 0.0087 & 0.0946 & 0.0125 & 0.3301 & 0.1221 & 0.0509 & 0.0187 & 0.2055 & 0.5391 & 0.0434 & 0.1289 \\
\midrule
\multirow{4}{*}{Llama-3.1-8B} 
  & PromptEOL & 0.0757 & 0.0045 & 0.0045 & 0.0044 & 0.0000 & 0.1515 & 0.0757 & 0.0099 & 0.0218 & 0.2012 & 0.0052 & 0.0009 & 0.0463 \\
  & Echo & 0.1035 & 0.0186 & 0.0245 & 0.2445 & 0.0239 & 0.1515 & 0.0757 & 0.2633 & 0.3693 & 0.0436 & 0.0044 & 0.0304 & 0.1128 \\
  & Token Prepending & 0.0757 & 0.0023 & 0.0023 & 0.0047 & 0.0059 & 0.1515 & 0.0757 & 0.0109 & 0.0109 & 0.1492 & 0.0029 & 0.0134 & 0.0421 \\
  & KV-Embedding & 0.2357 & 0.0100 & 0.0184 & 0.3367 & 0.0206 & 0.4842 & 0.1985 & 0.1755 & 0.0436 & 0.2658 & 0.6868 & 0.1535 & 0.2191 \\
\bottomrule
\end{tabular}%
}
\caption{LoCoV1 Experimental Results: 1024 Token Length}
\label{tab:loco_1024}
\end{table*}

\begin{table*}[htbp]
\centering
\resizebox{\textwidth}{!}{%
\begin{tabular}{llccccccccccccc}
\toprule
\multirow{2}{*}{\textbf{Model}} & \multirow{2}{*}{\textbf{Method}} & \textbf{2wiki} & \textbf{court} & \textbf{court} & \textbf{gov} & \textbf{legal} & \textbf{multi} & \textbf{pass} & \textbf{qasper} & \textbf{qasper} & \multirow{2}{*}{\textbf{qmsum}} & \textbf{stack} & \textbf{summ} & \multirow{2}{*}{\textbf{Avg.}} \\
& & \textbf{mqa} & \textbf{HTML} & \textbf{Text} & \textbf{rep} & \textbf{case} & \textbf{qa} & \textbf{ret} & \textbf{abs} & \textbf{title} & & \textbf{over} & \textbf{scrn} & \\
\midrule
\multirow{4}{*}{Mistral-7B} 
  & PromptEOL & 0.1308 & 0.0023 & 0.0023 & 0.0047 & 0.0106 & 0.1515 & 0.0599 & 0.0148 & 0.0000 & 0.0955 & 0.0032 & 0.0134 & 0.0407 \\
  & Echo & 0.1175 & 0.0051 & 0.0023 & 0.0047 & 0.0069 & 0.1973 & 0.0757 & 0.0109 & 0.0109 & 0.1181 & 0.0170 & 0.0160 & 0.0549 \\
  & Token Prepending & 0.0757 & 0.0028 & 0.0023 & 0.0047 & 0.0059 & 0.1515 & 0.0757 & 0.0109 & 0.0109 & 0.1492 & 0.0029 & 0.0134 & 0.0422 \\
  & KV-Embedding & 0.2246 & 0.0181 & 0.0197 & 0.1203 & 0.0266 & 0.4857 & 0.1307 & 0.0433 & 0.0415 & 0.3183 & 0.6072 & 0.1698 & 0.1838 \\
\midrule
\multirow{4}{*}{Qwen3-4B} 
  & PromptEOL & 0.0757 & 0.0015 & 0.0020 & 0.0073 & 0.0183 & 0.1515 & 0.0757 & 0.0070 & 0.0176 & 0.1492 & 0.0058 & 0.0134 & 0.0438 \\
  & Echo & 0.3004 & 0.0055 & 0.0038 & 0.0047 & 0.0059 & 0.1515 & 0.0572 & 0.0109 & 0.0617 & 0.1033 & 0.0289 & 0.0556 & 0.0658 \\
  & Token Prepending & 0.0757 & 0.0028 & 0.0023 & 0.0047 & 0.0059 & 0.1515 & 0.0757 & 0.0109 & 0.0109 & 0.1492 & 0.0029 & 0.0134 & 0.0422 \\
  & KV-Embedding & 0.1387 & 0.0123 & 0.0166 & 0.0299 & 0.0319 & 0.2505 & 0.1203 & 0.0505 & 0.0419 & 0.1741 & 0.5090 & 0.0542 & 0.1192 \\
\midrule
\multirow{4}{*}{Llama-3.1-8B} 
  & PromptEOL & 0.0551 & 0.0006 & 0.0023 & 0.0047 & 0.0059 & 0.1515 & 0.0757 & 0.0109 & 0.0218 & 0.1492 & 0.0005 & 0.0143 & 0.0410 \\
  & Echo & 0.0451 & 0.0012 & 0.0014 & 0.0058 & 0.0062 & 0.1515 & 0.0789 & 0.0171 & 0.0134 & 0.1035 & 0.0238 & 0.0141 & 0.0385 \\
  & Token Prepending & 0.0757 & 0.0028 & 0.0023 & 0.0047 & 0.0059 & 0.1515 & 0.0757 & 0.0109 & 0.0109 & 0.1492 & 0.0029 & 0.0134 & 0.0422 \\
  & KV-Embedding & 0.1591 & 0.0248 & 0.0355 & 0.1384 & 0.0552 & 0.4116 & 0.1237 & 0.0533 & 0.0433 & 0.3135 & 0.6733 & 0.1482 & 0.1817 \\
\bottomrule
\end{tabular}%
}
\caption{LoCoV1 Experimental Results: 2048 Token Length}
\label{tab:loco_2048}
\end{table*}

\begin{table*}[htbp]
\centering
\resizebox{\textwidth}{!}{%
\begin{tabular}{llccccccccccccc}
\toprule
\multirow{2}{*}{\textbf{Model}} & \multirow{2}{*}{\textbf{Method}} & \textbf{2wiki} & \textbf{court} & \textbf{court} & \textbf{gov} & \textbf{legal} & \textbf{multi} & \textbf{pass} & \textbf{qasper} & \textbf{qasper} & \multirow{2}{*}{\textbf{qmsum}} & \textbf{stack} & \textbf{summ} & \multirow{2}{*}{\textbf{Avg.}} \\
& & \textbf{mqa} & \textbf{HTML} & \textbf{Text} & \textbf{rep} & \textbf{case} & \textbf{qa} & \textbf{ret} & \textbf{abs} & \textbf{title} & & \textbf{over} & \textbf{scrn} & \\
\midrule
\multirow{4}{*}{Mistral-7B} 
  & PromptEOL & 0.0279 & 0.0023 & 0.0023 & 0.0051 & 0.0059 & 0.1816 & 0.1036 & 0.0133 & 0.0109 & 0.1649 & 0.0005 & 0.0123 & 0.0442 \\
  & Echo & 0.1432 & 0.0023 & 0.0147 & 0.0063 & 0.0149 & 0.2289 & 0.0920 & 0.0055 & 0.0109 & 0.1534 & 0.0228 & 0.0147 & 0.0591 \\
  & Token Prepending & 0.1097 & 0.0092 & 0.0133 & 0.0462 & 0.0709 & 0.2463 & 0.0771 & 0.1725 & 0.1611 & 0.1342 & 0.1327 & 0.0171 & 0.0992 \\
  & KV-Embedding & 0.1626 & 0.0310 & 0.0416 & 0.0789 & 0.0794 & 0.3861 & 0.1058 & 0.3443 & 0.3061 & 0.2569 & 0.5945 & 0.0947 & 0.2068 \\
\midrule
\multirow{4}{*}{Qwen3-4B} 
  & PromptEOL & 0.0868 & 0.0137 & 0.0201 & 0.0689 & 0.0390 & 0.2347 & 0.0816 & 0.2283 & 0.2197 & 0.1202 & 0.3995 & 0.0307 & 0.1286 \\
  & Echo & 0.1547 & 0.0096 & 0.0106 & 0.1195 & 0.0135 & 0.4423 & 0.1792 & 0.1291 & 0.0618 & 0.2233 & 0.1812 & 0.0564 & 0.1318 \\
  & Token Prepending & 0.0790 & 0.0094 & 0.0212 & 0.0666 & 0.0323 & 0.2608 & 0.0968 & 0.2468 & 0.1953 & 0.2239 & 0.2436 & 0.0297 & 0.1255 \\
  & KV-Embedding & 0.1118 & 0.0259 & 0.0369 & 0.0983 & 0.0964 & 0.3070 & 0.0952 & 0.4180 & 0.2973 & 0.1519 & 0.5003 & 0.0476 & 0.1822 \\
\midrule
\multirow{4}{*}{Llama-3.1-8B} 
  & PromptEOL & 0.1032 & 0.0121 & 0.0082 & 0.0852 & 0.0838 & 0.2628 & 0.0913 & 0.1678 & 0.1277 & 0.1383 & 0.4313 & 0.0306 & 0.1285 \\
  & Echo & 0.1713 & 0.0036 & 0.0023 & 0.0047 & 0.0051 & 0.1515 & 0.0872 & 0.0081 & 0.0078 & 0.1242 & 0.4377 & 0.4263 & 0.1191 \\
  & Token Prepending & 0.1233 & 0.0090 & 0.0084 & 0.0958 & 0.0579 & 0.2952 & 0.0989 & 0.2379 & 0.2285 & 0.1713 & 0.3247 & 0.0251 & 0.1397 \\
  & KV-Embedding & 0.1685 & 0.0503 & 0.0552 & 0.1138 & 0.1360 & 0.3872 & 0.1369 & 0.4459 & 0.3751 & 0.2862 & 0.6730 & 0.0567 & 0.2404 \\
\bottomrule
\end{tabular}%
}
\caption{LoCoV1 Experimental Results: 4096 Token Length}
\label{tab:loco_4096}
\end{table*}

\section{Attention Bias Experiment}\label{sec:attention_bias_detail}

During attention computation, we add a scalar bias $b$ to the pre-softmax logit when attending to the re-routed KV pair at position $0$. For a query $\mathbf{q}_i$ at any position $i$, the modified attention score is:
\begin{equation}
  \tilde{s}_{i,0} = \frac{\mathbf{q}_i^\top \mathbf{k}_0}{\sqrt{d}} + b
\end{equation}
The final attention weights $\alpha_{i,j}$ are then computed as:
\begin{equation}
  \alpha_{i,j} = \frac{\exp(\tilde{s}_{i,j})}{\sum_{k=0}^n \exp(\tilde{s}_{i,k})}
\end{equation}
where $\tilde{s}_{i,k} = s_{i,k}$ for all $k > 0$. A higher $b$ encourages stronger attention to the global summary.

Table~\ref{tab:attn_bias} shows the effect of varying $b$ on MTEB performance. The improvement from $b=0$ to $b=1.0$ confirms that the re-routed KV pairs provide useful context for embedding quality. The performance curve remains stable across a wide range, fluctuating within $\pm$1.5\% (0.519 to 0.534), suggesting that the injected KV pairs integrate naturally with the pre-trained attention mechanism. However, performance decreases when $b$ exceeds 3.0, indicating that excessive bias may cause over-reliance on the global summary at the expense of local details. We set $b=1.0$ to balance global context access with local specificity.

\begin{table*}[htbp]
\centering
\small
\begin{tabular}{ccccccccc}
\toprule
\textbf{Attention Bias} & \textbf{STS} & \textbf{Retr.} & \textbf{Class.} & \textbf{Pair.} & \textbf{Clust.} & \textbf{Rerank.} & \textbf{Summ.} & \textbf{Avg.} \\
\midrule
0.00 & 0.7690 & 0.2942 & 0.6948 & 0.7310 & 0.3855 & 0.5128 & 0.3096 & 0.5281 \\
1.00 & 0.7720 & 0.3014 & 0.6951 & 0.7564 & 0.3902 & 0.5145 & 0.3088 & 0.5341 \\
2.00 & 0.7736 & 0.3064 & 0.6949 & 0.7342 & 0.3926 & 0.5159 & 0.3107 & 0.5326 \\
3.00 & 0.7723 & 0.3089 & 0.6938 & 0.7329 & 0.3928 & 0.5135 & 0.3100 & 0.5320 \\
4.00 & 0.7649 & 0.3066 & 0.6907 & 0.7259 & 0.3928 & 0.5093 & 0.3027 & 0.5276 \\
5.00 & 0.7483 & 0.2952 & 0.6819 & 0.7078 & 0.3875 & 0.5010 & 0.3110 & 0.5190 \\
\bottomrule
\end{tabular}
\caption{Performance of Mistral-7B-Instruct with different attention bias values across various tasks.}\label{tab:attn_bias}
\end{table*}

\section{Geometry of the Embedding Space}\label{sec:geometry}

To evaluate the structural properties of the representation space, we utilize the alignment and uniformity metrics proposed by \citet{wang2020understanding}. The metrics are computed on a subset of 1,000 sentence pairs from the F2LLM dataset \citep{zhang2025f2llm} using Mistral-7B-Instruct. Let $f(x) \in \mathbb{R}^d$ be the $\ell_2$-normalized embedding of sentence $x$.

\paragraph{Alignment} 
Alignment evaluates the closeness of embeddings for semantically similar pairs. A lower alignment score indicates that the model effectively maps related instances to proximal regions in the latent space, preserving local semantic consistency:
\begin{equation}
\mathcal{L}_{\text{align}}(f; \alpha) \triangleq \mathbb{E}_{(x, y) \sim p_{\text{pos}}} \left[ \|f(x) - f(y)\|_2^\alpha \right]
\end{equation}
Following standard practice, we set $\alpha = 2$ in our experiments.

\paragraph{Uniformity} 
Uniformity assesses how evenly embeddings are distributed across the unit hypersphere. High uniformity (a more negative score) suggests that the embeddings capture maximal information and avoid the \textit{anisotropy} problem, where vectors collapse into a narrow cone:
\begin{equation}
\mathcal{L}_{\text{uniform}}(f; t) \triangleq \log \mathbb{E}_{x, y \stackrel{i.i.d.}{\sim} p_{\text{data}}} \left[ e^{-t \|f(x) - f(y)\|_2^2} \right]
\end{equation}
where $t=2$ is used for evaluation. Together, these metrics provide a comprehensive view of the representation's quality, balancing local clustering with global dispersion.

\section{Intrinsic Dimensionality Trajectories}\label{app:id_analysis}

\begin{figure}[t]
  \centering
  \includegraphics[width=0.9\linewidth]{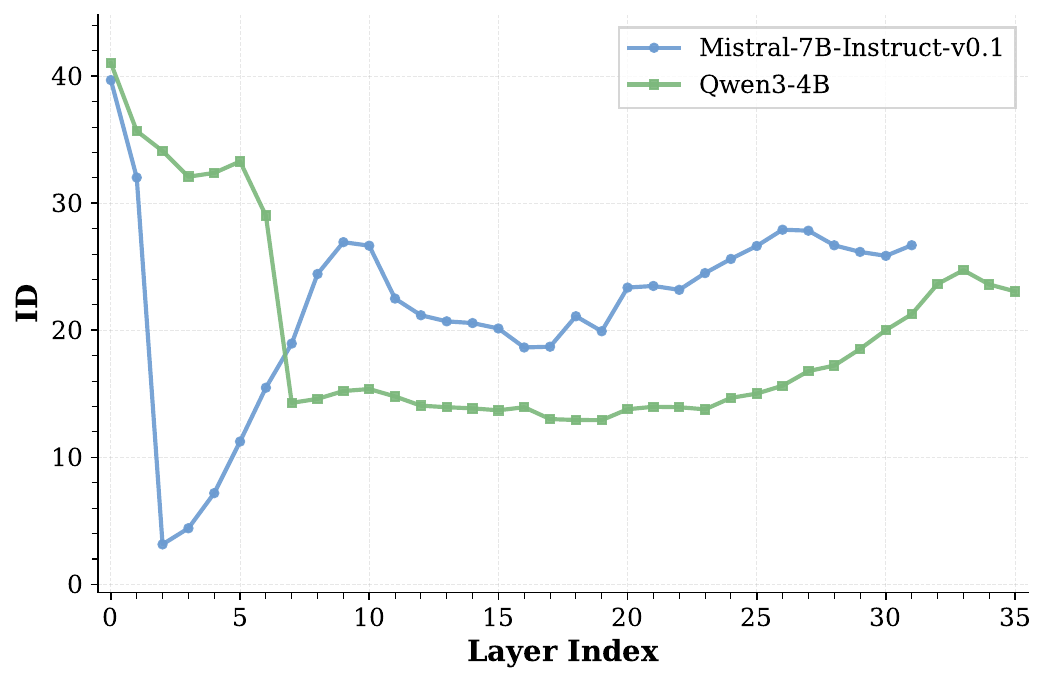}
  \caption{Intrinsic Dimensionality (ID) across layers for Mistral-7B-Instruct-v0.1 and Qwen3-4B. Different architectures exhibit distinct semantic compression patterns.}
  \label{fig:id_curves}
\end{figure}

Figure~\ref{fig:id_curves} presents the Intrinsic Dimensionality (ID) trajectories across all layers for Mistral-7B and Qwen3-4B. The ID is estimated using the TwoNN estimator~\citep{FaccoElena2017Etid} on 1,000 samples from the F2LLM dataset.

The visualization reveals distinct geometric patterns between the two architectures. Mistral-7B exhibits a sharp initial decline followed by fluctuations, with a stable low-ID region emerging around layers 13--19. Qwen3-4B demonstrates a more gradual descent, with its global minimum appearing around layers 17--19. In contrast, Qwen3-4B demonstrates a more gradual descent, with its global ID minimum appearing in the middle layers. Both models show a common trend where ID increases again as representations approach the final layers.

The variance in these trajectories indicates that the layers of maximal compression, where the representation manifold occupies the lowest number of intrinsic dimensions, differ by model. For Mistral-7B, the minimum ID occurs very early, whereas for Qwen3-4B, it is situated in the middle of the backbone. These empirical observations justify the use of an automated selection strategy over a fixed layer heuristic, as the ID-based approach adaptively identifies the specific layers where information is most condensed for a given architecture.

\section{Analysis of Layer Selection}\label{sec:layer_ablation}

To validate the efficacy of Intrinsic Dimensionality (ID) for layer selection, we compare our strategy against uniform partitioning strategies: Early, Middle, and Late layers, each covering approximately one-third of the total model depth. As summarized in Table~\ref{tab:layer_selection_ablation}, re-routing layers $\mathcal{L}$ derived from ID analysis consistently yield the highest average performance while maintaining a significantly more parsimonious selection of layers.

On Qwen3-4B, re-routing through early layers (0--11) results in the lowest average score (0.409). This supports the premise that shallow layers focus on surface-level lexical and syntactic features, which lack the semantic density required for effective global information redistribution. Conversely, ID selection identifies a specific range of layers (12--21) that achieves an average score of 0.494, outperforming the broader middle-third strategy (12--23).

A similar pattern emerges on Mistral-7B-Instruct-v0.1. On Mistral-7B, our ID-based selection utilizes only 7 layers (13--19) yet surpasses the 10-layer middle strategy (10--19). This efficiency indicates that minimum intrinsic dimensionality accurately pinpoints the layer range where the representation manifold achieves maximal compression. By injecting global context at these specific positions, our method avoids the noisy low-level features of early layers and the task-specific prediction bias inherent in final-layer hidden states. These findings suggest that semantic redistribution is most effective when applied at the stage of peak abstraction, rather than through uniform or heuristic selection.

\section{Impact of Attention Constraints}\label{sec:attention_ablation}

This section provides additional analysis on why naive mask removal fails, complementing the results in Table~\ref{tab:bi_attn_main}.

Decoder-only LLMs are pre-trained under strict causal constraints, where each token only attends to preceding positions. When bidirectional attention is forced on a frozen model, every token suddenly receives key-value states from future positions that were never encountered during pre-training. These out-of-distribution states disrupt the learned attention patterns, causing the model to produce degraded representations.

The consistent performance collapse across both Qwen3-4B and Mistral-7B (Table~\ref{tab:bi_attn_main}) confirms that this is not a model-specific issue but a fundamental incompatibility between bidirectional attention and causally pre-trained weights. Notably, the degradation is most severe on Retrieval tasks, where Bi-Attn scores drop to near-zero (0.02--0.03), suggesting that document-level semantics are particularly disrupted.

KV-Embedding avoids this issue by operating within the causal framework. The re-routed KV states originate from the final token, which has legitimately attended to all preceding positions during the forward pass. By prepending these states rather than altering the attention mask, we preserve the model's internal dynamics while enabling global context access.


\section{Prompt Design}\label{sec:prompt_ablation}

Table~\ref{tab:prompt} evaluates the impact of five different prompt variations on the performance of Mistral-7B-Instruct-v0.1. The results indicate that the model's performance is consistent across different instruction phrasings, with the average scores ranging narrowly from 0.529 to 0.540. This stability suggests that the internal re-routing mechanism is robust to the specific choice of natural language instructions.

While the overall variance is minimal, specific templates exhibit localized advantages. For example, Prompt 0 (``Compress the context in one word''), which serves as our default configuration, achieves the highest scores in STS (0.772), Classification (0.695), and Summarization (0.309). This indicates its superior ability to capture global semantic information. On the other hand, Prompt 2 (``Extract key concept'') shows a slight edge in Retrieval (0.336) and Clustering (0.419), suggesting that phrasings emphasizing extraction may better align with task-specific representations for dense vector spaces.

\section{Pooling Strategy}\label{sec:pooling_ablation}

Table~\ref{tab:pooling_ablation} compares pooling strategies after KV re-routing. Mean pooling consistently performs worst across all backbones. This likely reflects that early tokens retain lexical noise that averaging cannot filter, leading to semantic dilution even with re-routed context.

Hybrid pooling (averaging the last-token and mean representations) achieves the highest average scores. This strategy combines the distributive evidence across the sequence with the globally-informed summary captured by the final state.

\begin{table*}[htbp]
\centering
\small
\begin{tabular}{llcccccccc}
\toprule
\textbf{Model} & \textbf{Layer Selection} & \textbf{STS} & \textbf{Retr.} & \textbf{Class.} & \textbf{Pair.} & \textbf{Clust.} & \textbf{Rerank.} & \textbf{Summ.} & \textbf{Avg.} \\
\midrule
\multirow{4}{*}{Qwen3-4B} 
& Early Layers (0--11) & 0.527 & 0.253 & 0.578 & 0.470 & 0.362 & 0.441 & 0.234 & 0.409 \\
& Middle Layers (12--23) & 0.713 & \textbf{0.284} & \textbf{0.639} & 0.677 & 0.391 & 0.490 & 0.255 & 0.493 \\
& Late Layers (24--35) & \textbf{0.726} & 0.272 & 0.634 & 0.666 & \textbf{0.399} & 0.477 & 0.247 & 0.489 \\
& ID Selected (12--21) & 0.714 & 0.277 & 0.638 & \textbf{0.680} & 0.390 & \textbf{0.501} & \textbf{0.257} & \textbf{0.494} \\
\midrule
\multirow{4}{*}{\shortstack{Mistral-7B-\\Instruct-v0.1}} 
& Early Layers (0--9) & 0.759 & 0.277 & 0.691 & 0.732 & 0.381 & 0.505 & 0.301 & 0.521 \\
& Middle Layers (10--19) & 0.771 & 0.301 & 0.695 & 0.731 & 0.389 & 0.514 & 0.304 & 0.529 \\
& Late Layers (20--31) & 0.771 & 0.297 & \textbf{0.696} & 0.735 & 0.382 & 0.510 & 0.300 & 0.527 \\
& ID Selected (13--19) & \textbf{0.772} & \textbf{0.301} & 0.695 & \textbf{0.756} & \textbf{0.390} & \textbf{0.515} & \textbf{0.309} & \textbf{0.534} \\
\bottomrule
\end{tabular}%
\caption{Ablation study on layer selection. We compare uniform strategies (Early, Middle, Late) with ID-based selection (layer ranges shown in parentheses). ID selection achieves the best average performance while using fewer layers. }\label{tab:layer_selection_ablation}
\end{table*}

\begin{table*}[htbp]
\centering
\small
\begin{tabular}{clcccccccc}
\toprule
\textbf{ID} & \textbf{Prompt Template} & \textbf{STS} & \textbf{Retr.} & \textbf{Class.} & \textbf{Pair.} & \textbf{Clust.} & \textbf{Rerank.} & \textbf{Summ.} & \textbf{Avg.} \\
\midrule
0 & Compress the context in one word & \textbf{0.772} & 0.301 & \textbf{0.695} & 0.756 & 0.390 & 0.515 & \textbf{0.309} & 0.534 \\
1 & Summarize in one word & 0.768 & 0.292 & 0.683 & 0.756 & 0.391 & 0.518 & 0.303 & 0.530 \\
2 & Extract key concept & 0.754 & \textbf{0.336} & 0.666 & \textbf{0.775} & \textbf{0.419} & \textbf{0.533} & 0.299 & \textbf{0.540} \\
3 & Represent in one word & 0.766 & 0.294 & 0.683 & 0.755 & 0.392 & 0.517 & 0.299 & 0.529 \\
4 & Compress to one word & 0.767 & 0.294 & 0.684 & 0.766 & 0.393 & 0.518 & 0.296 & 0.531 \\
\bottomrule
\end{tabular}
\caption{Ablation on prompt templates (Mistral-7B). Performance is stable across prompts, with ``Extract key concept'' showing strength on retrieval and clustering tasks.}\label{tab:prompt}
\end{table*}

\begin{table*}[htbp]
\centering
\small
\begin{tabular}{llcccccccc}
\toprule
\textbf{Model} & \textbf{Pooling Strategy} & \textbf{STS} & \textbf{Retr.} & \textbf{Class.} & \textbf{Pair.} & \textbf{Clust.} & \textbf{Rerank.} & \textbf{Summ.} & \textbf{Avg.} \\
\midrule
\multirow{3}{*}{Qwen3-4B} 
& Last Token & \textbf{0.718} & \underline{0.266} & \textbf{0.652} & \underline{0.638} & \textbf{0.397} & \textbf{0.511} & \textbf{0.271} & \underline{0.493} \\
& Mean Pooling & 0.504 & 0.177 & 0.516 & 0.535 & 0.351 & 0.411 & 0.219 & 0.388 \\
& Hybrid Pooling & \underline{0.714} & \textbf{0.277} & \underline{0.638} & \textbf{0.680} & \underline{0.390} & \underline{0.501} & \underline{0.257} & \textbf{0.494} \\
\midrule
\multirow{3}{*}{\shortstack{Mistral-7B-\\Instruct-v0.1}} 
& Last Token & \underline{0.767} & \underline{0.253} & \underline{0.694} & \underline{0.683} & 0.338 & \underline{0.503} & \underline{0.292} & \underline{0.504} \\
& Mean Pooling & 0.588 & 0.226 & 0.566 & 0.644 & \underline{0.379} & 0.446 & 0.260 & 0.444 \\
& Hybrid Pooling & \textbf{0.772} & \textbf{0.301} & \textbf{0.695} & \textbf{0.756} & \textbf{0.390} & \textbf{0.515} & \textbf{0.309} & \textbf{0.534} \\
\bottomrule
\end{tabular}%
\caption{Ablation on pooling strategies. Hybrid pooling (last-token + mean) achieves the best overall performance. \textbf{Bold}: best, \underline{underline}: second-best.}\label{tab:pooling_ablation}
\end{table*}

\end{document}